# Balancing multiscale similarity and cartographic constraints: A similarity-driven optimization framework for line generalization

Pengbo Li [a,b,c], Haowen Yan [a,b,c,*], Xiaomin Lu [a,b,c], Binbin Lin [a,b,c]

[a]*Faculty of Geomatics, Lanzhou Jiaotong University, Lanzhou 730070, China;*
[b]*National-Local Joint Engineering Research Center of Technologies and Applications for National Geographic State Monitoring, Lanzhou 730070, China;*
[c]*Key Laboratory of Science and Technology in Surveying & Mapping, Lanzhou 730070, China*

# Balancing multiscale similarity and cartographic constraints: A similarity-driven optimization framework for line generalization

## Abstract

Cartographic generalization is essential for generating multiscale map representations by balancing information preservation and cartographic readability. However, automated generalization remains challenging because existing approaches often treat spatial similarity evaluation, cartographic constraints, and parameter optimization as separate processes, limiting adaptive and interpretable control across scales. This study formulates cartographic generalization as a constrained multiscale similarity optimization problem and proposes a similarity-driven framework for adaptive generalization control. The framework integrates multiscale spatial similarity as an optimization objective to quantify representation consistency between original and generalized data, while incorporating cartographic constraints to regulate readability, smoothness, and geometric validity. A unified objective function is optimized to automatically identify scale-dependent parameter configurations for different generalization algorithms. Experiments using multiple line simplification algorithms, target scales, and similarity measures, including geometric, structural, and learning-based metrics, demonstrate that the proposed framework achieves an effective balance between similarity preservation and cartographic abstraction. The results further show that combining similarity optimization with cartographic constraints provides more consistent and interpretable parameter control than relying on similarity evaluation alone. This study provides a unified optimization perspective that connects similarity assessment, constraint modeling, and algorithm control, contributing to adaptive and automated cartographic generalization.

## 1. Introduction

Cartographic generalization is fundamental to multiscale spatial representation, enabling geographic information to be transformed into map products appropriate for different scales, purposes, and levels of detail (Yan et al. 2025). As national mapping agencies and web mapping platforms increasingly rely on automated production workflows, modern mapping systems require not only efficient geometric simplification, but also adaptive control mechanisms capable of balancing information preservation and cartographic readability across varying target scales. Although considerable progress has been achieved through rule-based methods, constraint-based models, and more recently machine learning approaches (Zhou et al. 2026), automated generalization remains difficult to control in a consistent and interpretable manner. In particular, the relationships among algorithm parameters, target scales, and resulting cartographic quality are still insufficiently formalized, limiting the development of automated multiscale mapping systems.

Existing studies have approached automated generalization from several complementary perspectives. Early research primarily relied on cartographic rules and heuristic workflows to organize generalization operations according to expert knowledge and predefined mapping principles. Constraint-based approaches later formalized cartographic requirements such as legibility, minimum distance, and shape preservation into optimization frameworks (Harrie 1999, Sester 2005), improving the consistency and automation of generalization processes. More recent advances in deep learning have enabled data-driven prediction of generalization operations and parameter configurations from training examples. Despite these developments, existing optimization frameworks predominantly optimize algorithm-specific geometric objectives or weighted combinations of cartographic constraints, while spatial similarity, although widely adopted for quality assessment, is rarely embedded as a direct optimization objective. This separation between evaluation and control hinders the development of a unified optimization strategy that can be consistently applied across different generalization algorithms.

One major challenge in cartographic generalization is how to effectively control algorithm behavior across different map scales. In practical map production, experienced cartographers typically refine generalization parameters through iterative comparisons between original and generalized representations (Figure 1), considering differences in geometric structures, visual patterns, and overall cartographic quality. This iterative process indicates that parameter adjustment is implicitly guided by comparisons between multiscale representations. Therefore, a key research question arises: can such similarity-based evaluation be quantitatively modeled and incorporated into the generalization process to enable automatic parameter optimization? This question motivates the exploration of spatial similarity as a potential optimization objective for automated generalization. By integrating multiscale similarity with cartographic constraints, generalization parameters can be optimized according to both information preservation and cartographic requirements. This perspective provides a foundation for developing automated generalization systems by connecting quality assessment, optimization, and algorithm control within a unified framework.

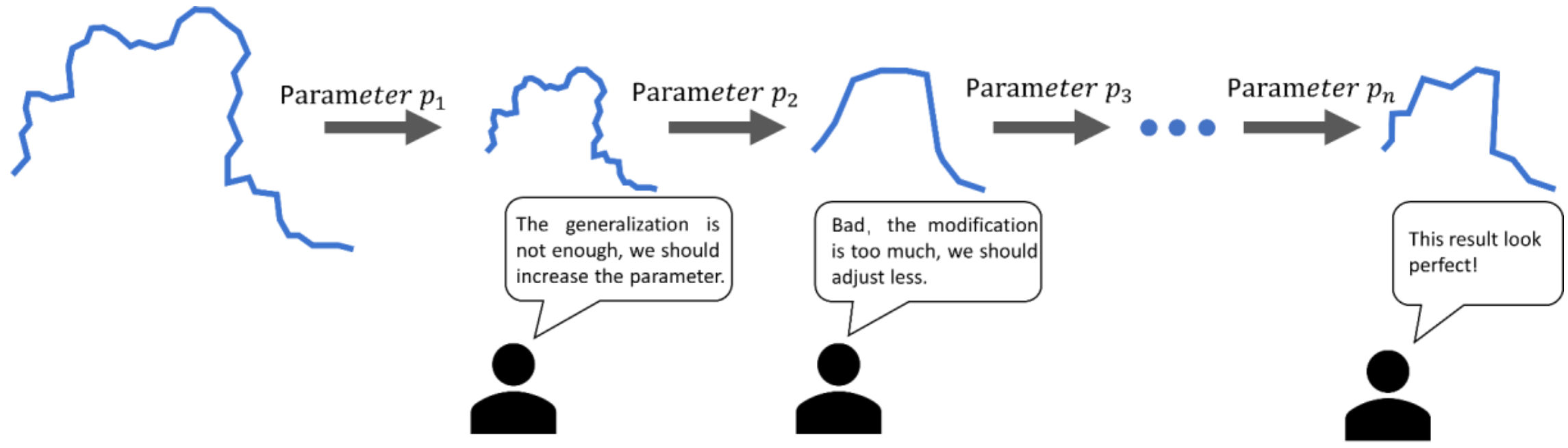

Figure 1. Human-involved parameter optimization of line simplification algorithm

Spatial similarity provides a natural basis for bridging optimization and quality assessment, as generalized representations are expected to preserve the essential structure of original data across scales (Mackaness & Ruas 2007). Various geometric, topological, and structural measures have been proposed to evaluate such consistency, with recent learning-based models further capturing high-level perceptual characteristics (Li et al. 2023). These developments provide the technical foundation for incorporating similarity into optimization frameworks, allowing similarity to function as an objective that guides generalization rather than merely assessing its outcomes. In this view, cartographic generalization can be formulated as a constrained optimization problem, where similarity quantifies information retention and cartographic constraints regulate visual quality and readability.

To address the aforementioned limitations, we reformulate cartographic generalization as a constrained similarity-driven optimization problem. Under this formulation, generalization is no longer treated merely as a sequence of heuristic geometric operations, but as an optimization process seeking to balance information preservation and cartographic abstraction across scales. Multiscale spatial similarity is modeled as the primary optimization objective describing the consistency between original and generalized representations, while cartographic requirements are represented as quantitative constraints governing readability, visual quality, and geometric validity.

Based on this perspective, we propose a similarity-driven optimization framework that integrates similarity preservation and constraint satisfaction within a unified objective function for adaptive and scale-aware control of generalization algorithms. The framework consists of three major components: (1) multiscale spatial similarity measurement, (2) quantitative modeling of cartographic constraints, and (3) optimization-driven parameter selection. Using line simplification as a representative case, experiments are conducted across multiple algorithms and target scales to evaluate the effectiveness of the proposed framework.

The main methodological contributions of this framework are threefold. First, it repositions spatial similarity from a passive post-evaluation metric to an active optimization objective that directly guides parameter selection in the generalization process, rather than merely serving as an additional evaluation criterion. Second, it establishes a unified objective function that integrates similarity preservation with multiple cartographic constraints, including clarity, smoothness, and topological validity, thereby bridging the conventional separation between quality assessment and optimization. Third, it provides a transferable parameter optimization strategy applicable to multiple line simplification algorithms, enabling consistent and adaptive parameter selection within a unified optimization framework.

## 2. Related works

### *2.1. Constraints for automating map generalization*

Constraint-based modeling has been one of the dominant paradigms for automating cartographic generalization. Early studies translated cartographic principles into operational rules that guided generalization operations according to expert knowledge and predefined mapping requirements (Beard 1991; Weibel 1996; Weibel & Dutton 1998). Harrie (1999) further formalized these principles through a constraint satisfaction framework, where cartographic requirements such as minimum distance, shape preservation, and legibility were explicitly represented and iteratively resolved. Subsequent studies extended this framework by distinguishing internal constraints related to individual object characteristics from relational constraints describing interactions among multiple geographic features (Burghardt et al. 2007; Sayidov et al. 2020; Li et al. 2024). More recent efforts have incorporated optimization techniques and machine learning models to improve the adaptability of constraint handling under different geographic contexts and target scales (Courtial et al. 2022; Courtial et al. 2024). However, existing constraint-based approaches mainly focus on satisfying predefined cartographic requirements, while the quantitative relationship between constraint satisfaction, information preservation, and algorithm parameter adjustment remains insufficiently modeled. Consequently, constraints alone provide limited guidance for explicitly optimizing information preservation while adapting parameter selection during multiscale generalization.

### *2.2. Spatial similarity in map generalization*

Spatial similarity represents a fundamental criterion for evaluating the quality of generalized representations, as effective generalization should preserve essential geometric, structural, and perceptual characteristics of original geographic features across scales (Mackaness & Ruas 2007; Yan 2024). Existing studies have developed various similarity measures based on geometric displacement, shape characteristics, topology preservation, and spatial patterns to quantitatively assess the consistency between original and generalized representations (Ai et al. 2015; Li et al. 2023; Liu et al. 2026). With increasing attention to multiscale representation, recent studies have emphasized that similarity should be evaluated across multiple levels of detail rather than at a single target scale (Wang et al. 2024). Learning-based similarity models have further demonstrated the capability of capturing complex geometric and perceptual characteristics beyond traditional handcrafted measures (Li et al. 2023). Despite these advances, similarity has primarily remained an evaluation-oriented concept used for assessing generalization results after processing. Its potential role as an optimization objective that actively guides parameter selection during algorithm execution has not been sufficiently explored.

### *2.3. Optimization and control of generalization*

Process control in map generalization focuses on determining appropriate operations, parameter settings, and execution strategies to achieve desired cartographic outcomes. Early approaches mainly relied on rule-based workflows, where experts predefined operation sequences according to cartographic knowledge and production experience (Barbara et al. 1991). However, such static workflows lack flexibility when dealing with diverse geographic configurations and varying scale requirements.

To improve adaptability, goal-oriented and constraint-driven approaches introduced dynamic selection of generalization operations based on feature characteristics, contextual relationships, and constraint satisfaction (Sester 2005; Ruas & Duchêne 2007; Jin et al. 2024). Agent-based models further enhanced automation by allowing geographic objects to negotiate generalization actions through local interactions and behavioral rules (Duchêne et al. 2018). Beyond operation selection, heuristic optimization methods such as genetic algorithms, simulated annealing, and evolutionary strategies have been widely employed to search for suitable parameter configurations under predefined objectives and constraints. However, these optimization approaches are highly dependent on predefined objective functions, which are often designed separately from similarity-based quality assessment.

Recent advances in artificial intelligence and machine learning have enabled data-driven prediction of operation sequences and parameter configurations from existing examples (Du et al. 2022; Yu & Chen 2022; Courtial et al. 2023; Yan & Yang 2024). Nevertheless, most existing approaches focus on selecting operations or predicting parameters, while lacking a unified optimization mechanism that quantitatively connects similarity-based quality objectives, cartographic constraints, and parameter optimization across scales.

### *2.4. Research gap and motivation*

Despite substantial progress in constraint modeling, similarity evaluation, and optimization-based automation, three interrelated limitations remain. First, constraint-based approaches mainly describe cartographic requirements but provide limited mechanisms for explicitly optimizing information preservation while adapting parameter selection across scales. Second, similarity measures are predominantly employed for post-processing evaluation rather than as active objectives guiding algorithm execution and parameter selection. Third, existing optimization frameworks lack a unified formulation that integrates cartographic constraints, multiscale similarity, and parameter optimization within a coherent mathematical formulation.

These limitations motivate the formulation of the generalization process as a constrained similarity-driven optimization problem, where similarity serves as the optimization objective and cartographic constraints regulate acceptable transformations. Accordingly, this study proposes a similarity-driven optimization framework integrating multiscale similarity measurement, quantitative constraint modeling, and optimization-based parameter selection for adaptive generalization control.

## 3. Methodology

### *3.1 Problem formulation: Generalization as constrained similarity optimization*

Cartographic generalization is traditionally considered as a sequence of geometric operations that reduce spatial detail to generate map representations suitable for smaller scales. In this study, it is reformulated as a constrained multiscale representation transformation problem, where the objective is to determine an optimal transformation that balances information preservation and cartographic abstraction. Specifically, a generalized representation should preserve essential geometric and structural characteristics of the source data while satisfying cartographic requirements at the target scale, including legibility, smoothness, and geometric validity.

Spatial similarity provides a quantitative basis for modeling such multiscale transformations. It measures the degree of structural consistency retained between source and generalized representations and therefore reflects the amount of information preserved during scale transformation. Unlike conventional approaches that employ similarity only as a post-processing evaluation measure, this study incorporates similarity as an explicit optimization objective that guides the selection of generalization parameters. However, similarity preservation should be balanced with necessary cartographic abstraction: excessive preservation may constrain appropriate simplification, whereas excessive transformation may introduce structural distortion and reduce cross-scale consistency.

Based on this perspective, cartographic generalization is formulated as a parameterized transformation between multiscale spatial representations. Let $X_s$ denote the spatial representation at the source scale and $X_t$ denote the desired representation at the target scale. The generalization process can be expressed as:

$$f_\theta : X_s \rightarrow X_t \tag{1}$$

where $\theta$ denotes the parameters that determine the transformation process. The objective of generalization is therefore to determine an optimal parameter configuration $\theta^*$ that maximizes similarity preservation while satisfying cartographic constraints associated with the target-scale representation.

This formulation transforms cartographic generalization from an algorithm-specific geometric operation into a constrained optimization problem, providing a unified interpretation that connects multiscale representation transformation, similarity preservation, and adaptive parameter optimization.

### *3.2. Similarity-driven optimization framework*

Building on the above formulation, this study develops a similarity-driven optimization framework for adaptive control of cartographic generalization. Instead of manually specifying algorithm parameters for different target scales, the proposed framework automatically searches parameter configurations by jointly considering multiscale similarity preservation and cartographic constraint satisfaction. As illustrated in Figure 2, the framework consists of three interconnected components: (1) multiscale similarity measurement, (2) quantitative cartographic constraint modeling, and (3) optimization-based parameter selection.

The first component, multiscale similarity measurement, evaluates the degree of information preservation and structural consistency between original and generalized representations. In contrast to conventional similarity measures used only for post-generalization evaluation, the proposed framework incorporates similarity as an explicit optimization objective and a quantitative feedback signal for determining appropriate transformation intensity across scales.

The second component, cartographic constraint modeling, converts cartographic requirements, including legibility, geometric validity, smoothness, and visual quality, into quantitative constraint functions. These constraints define the feasible solution space and regulate the degree of abstraction required for target-scale representation.

The third component, optimization-based parameter selection, searches the parameter space of generalization algorithms to maximize the integrated objective function composed of similarity preservation and constraint satisfaction. During generalization, similarity preservation and cartographic constraints may exhibit competing tendencies: excessive simplification may reduce structural consistency, whereas excessive preservation may restrict necessary abstraction. Therefore, the optimal parameter configuration is regarded as the solution that achieves the best balance between multiscale representation consistency and cartographic readability.

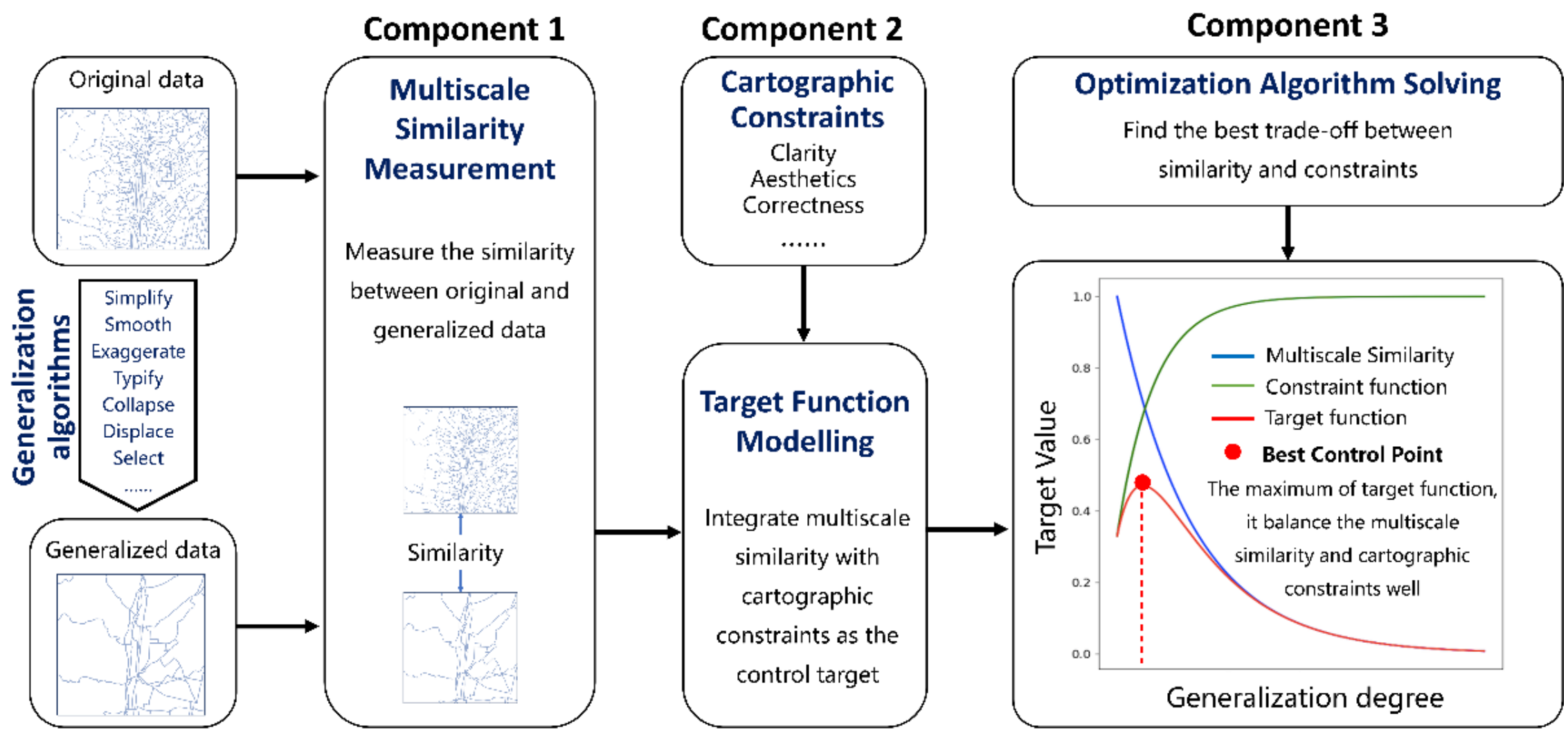

Figure 2. Similarity-driven optimization framework for adaptive generalization control

To further formalize the proposed framework, line generalization is selected as a representative implementation case. Let $l$ denote an original line object and $l_g$ denote its generalized counterpart. The original and target map scales are denoted as $1{:}S_o$ and $1\!:\!S_t\ (S_t \geq S_o)$, respectively. A generalization algorithm $A(\cdot)$ produces the generalized result as

$$l_g = A(l, \boldsymbol{x}) \tag{2}$$

where $\boldsymbol{x}$ represents one or more parameters controlling the generalization process.

A multiscale spatial similarity metric $Sim(\cdot)$ is introduced to evaluate information preservation between $l$ and $l_g$, where $0 \leq Sim(l, l_g) \leq 1$. Higher similarity values indicate stronger preservation of spatial characteristics across scales.

Meanwhile, a set of constraint satisfaction functions $\{C_i(\cdot) \mid i = 1, \dots, k\}$ is defined to quantify the fulfillment of cartographic requirements, where $0 \leq C_i(\cdot) \leq 1$. The individual constraints are aggregated into an overall constraint function $C_{total}(\cdot) = f(C_1, C_2, \dots, C_k)$ according to the requirements of the generalization task, $f(\cdot)$ is a mapping function to combine multiple constraints. Based on the similarity objective and constraint satisfaction, the integrated optimization objective is defined as:

$$T(\mathbf{x}) = g\left(Sim\big(l, A(l, \mathbf{x})\big), C_{total}\big(A(l, \mathbf{x})\big)\right) \tag{3}$$

where $Sim(\cdot)$ evaluates information preservation and $C_{\text{total}}(\cdot)$ measures the satisfaction of cartographic constraints in terms of clarity, aesthetics, and data regularity, $g(\cdot)$ combines similarity preservation and cartographic constraint satisfaction. The optimal parameter configuration is obtained by:

$$\boldsymbol{x}^* = \arg\max_{\boldsymbol{x}} T(\boldsymbol{x}) \tag{4}$$

The corresponding generalized result $l_g^*$ represents the optimal line generalization outcome at the target scale given the algorithm.

$$l_g^* = A(l, \boldsymbol{x}^*) \tag{5}$$

Under this formulation, the optimal generalized representation is not defined by a single geometric criterion, but by the parameter configuration that achieves the optimal trade-off between multiscale representation consistency and cartographic requirements.

### *3.3. Multiscale similarity measures for line generalization*

Multiscale spatial similarity serves as the primary optimization objective in the proposed framework, as it quantitatively characterizes the degree of representation consistency between original and generalized line features across different scales. Unlike conventional applications where similarity measures are only used for evaluating generalization quality after processing, this study incorporates similarity directly into the optimization process to guide parameter selection.

To capture both geometric fidelity and higher-level structural consistency of linear representations, a deep metric learning (DML) based similarity model is adopted as the primary similarity measure. Two classical geometric similarity metrics, Median Hausdorff Distance (MHD) and Fréchet Distance (FD), are further introduced as baseline measures for comparative analysis.

Specifically, this study adopts the Siamese LineStringNet architecture proposed by Li et al. (2023) to learn scale-aware latent representations of line features. The model is pre-trained using the publicly available dataset released in the original study, providing a consistent and reproducible basis for similarity computation. As illustrated in Figure 3, the network maps a line feature into a continuous embedding space, where the distance between embeddings reflects structural dissimilarity.

Given a line feature $l$, the embedding function $f_\theta(\cdot)$ parameterized by $\theta$ maps the input line into a feature vector:

$$\boldsymbol{v} = f_\theta(l) \tag{6}$$

The similarity between the original line $l_o$ and the generalized line $l_g$ is then computed as the distance between their embeddings:

$$d_{DML}(l_o, l_g) = \| f_\theta(l_o) - f_\theta(l_g) \| \tag{7}$$

Compared with handcrafted geometric measures, the DML-based similarity model learns representation features directly from multiscale line examples and can capture complex spatial characteristics, including global shape configuration, curvature patterns, and perceptual structural consistency. Therefore, it provides a more adaptive measurement of multiscale representation similarity for similarity-driven generalization control. To evaluate the effectiveness of the learned similarity representation, two widely used geometric similarity measures are employed as comparison baselines.

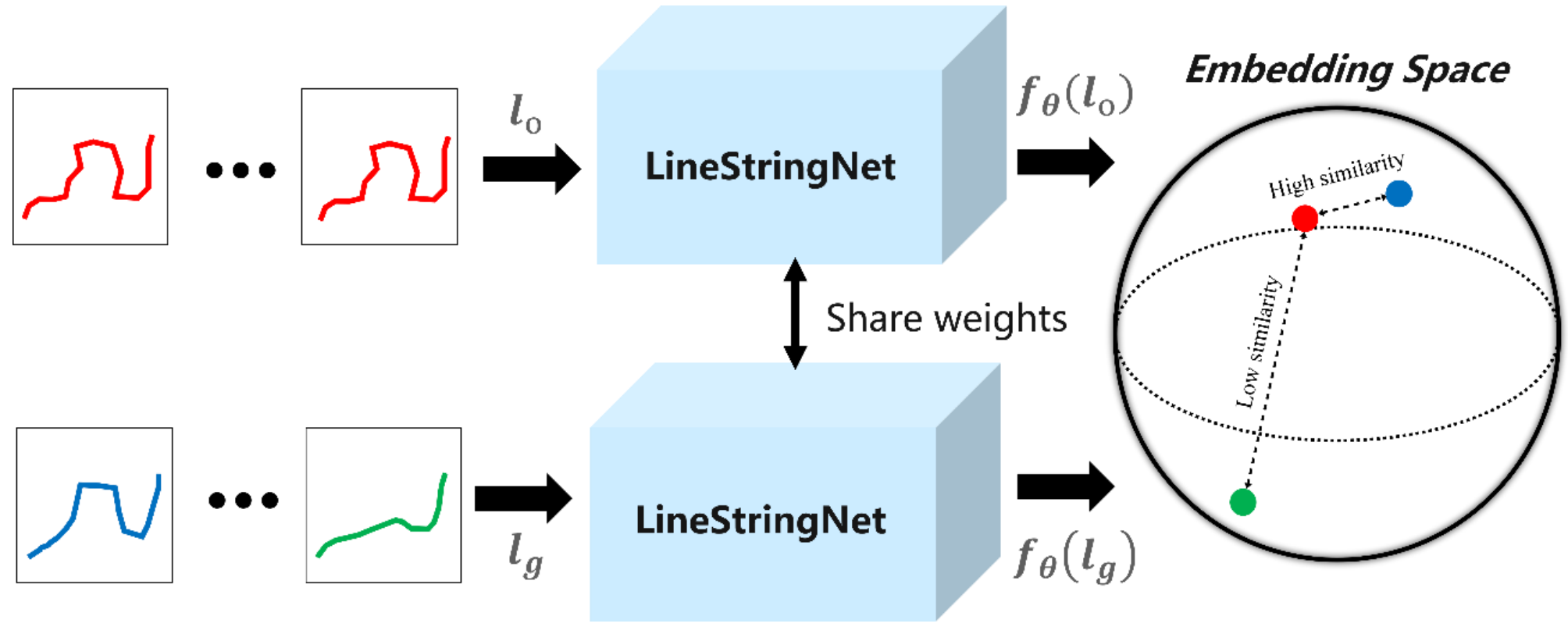

Figure 3. Deep metric learning framework for line similarity measure

The Median Hausdorff Distance (MHD) measures positional discrepancy between two line representations while reducing sensitivity to local outliers through median aggregation (Tong et al., 2014). Lower MHD values indicate smaller positional deviations between original and generalized lines:

$$d_{MHD}(l_o, l_g) = \mathrm{median}_{p \in l_o}\left(\min_{q \in l_g} \|p - q\|\right) \tag{8}$$

The Fréchet distance accounts for both spatial proximity and the ordering of points along the curves (Yu & Chen 2022). If the original and generalized lines are parameterized as $l_o(t)$ and $l_g(t)$, the Fréchet distance is defined as:

$$d_{FD}(l_o, l_g) = \inf_{\alpha,\beta} max_{t \in [0,1]} \| l_o(\alpha(t)) - l_g(\beta(t)) \| \tag{9}$$

where $\alpha$ and $\beta$ are continuous, non-decreasing reparameterizations of the curves. Compared to Hausdorff distance, the Fréchet distance is more sensitive to the shape distortions such as bending, stretching, and local misalignment.

Within the proposed framework, the DML-based similarity measure is adopted as the optimization objective for parameter selection, while MHD and FD serve as geometric baselines for evaluating representation consistency. This combination enables the framework to consider both explicit geometric fidelity and implicit structural similarity during multiscale line generalization.

### *3.4. Cartographic constraints modelling for line generalization*

In the proposed framework, cartographic constraints are formulated as quantitative satisfaction functions to regulate the acceptable abstraction level of generalized line representations. Unlike conventional constraint-based approaches that treat constraints mainly as independent requirements, this study integrates constraint satisfaction directly into the optimization objective to jointly balance representation similarity and cartographic quality. Three types of cartographic requirements are considered: clarity, aesthetic quality, and geometric validity.

(1) Clarity constraint: Clarity constraints ensure that generalized linear features remain visually distinguishable at the target scale. During scale reduction, spatial distances between nearby parts of a line may become smaller than the minimum perceptible size, resulting in visual conflicts and reduced readability. Therefore, clarity is modeled according to the minimum visible dimension principle proposed by Li and Openshaw (1992). The minimum visible distance at the target scale is defined as:

$$D_{min} = S_t \cdot D \tag{10}$$

where $D$ is the minimum visible size, representing the smallest distinguishable value on the target map, commonly set to 0.04 mm, and $S_t$ is the target scale.

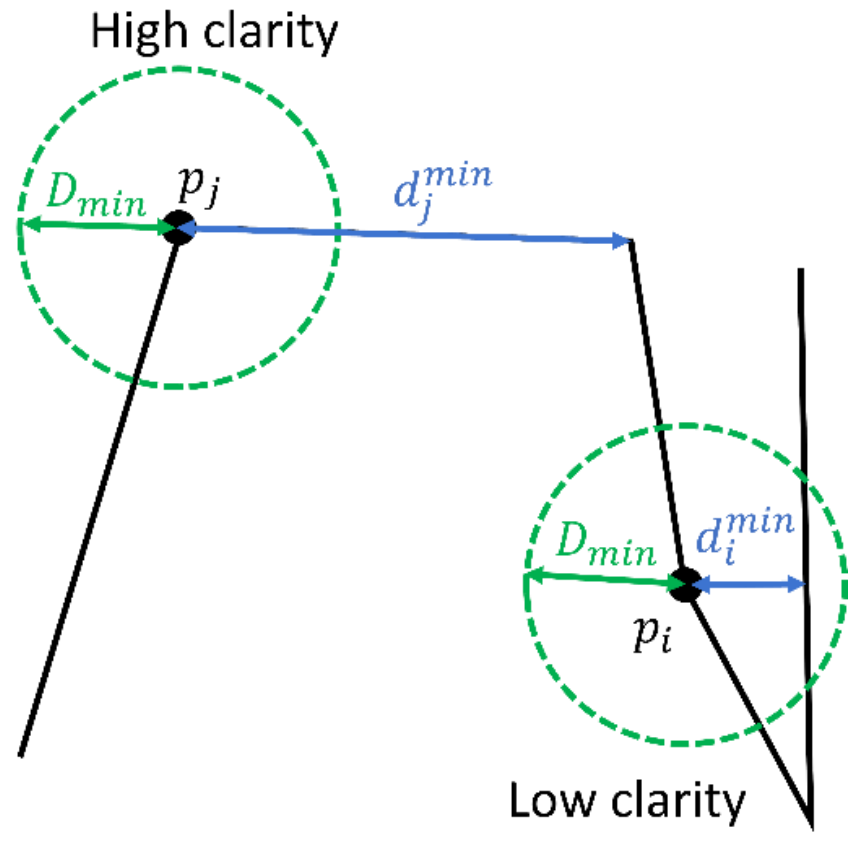

Figure 4. Schematic diagram showing the calculation for vertex clarity

For a vertex $p_i$, the minimum distance between this vertex and non-adjacent segments of the same line is calculated as $d_i^{min}$. If $d_i^{\min} > D_{\min}$, the vertex can be clearly distinguished from other parts of the line at the target scale, indicating high clarity (Figure 4). Otherwise, the vertex is prone to visual confusion and exhibits low clarity. A vertex-level clarity satisfaction score is defined as:

$$Clarity_i = \min\left(\frac{d_i^{min}}{D_{min}}, 1\right) \tag{11}$$

where values close to 1 indicate that the vertex satisfies the visibility requirement. The overall clarity constraint is calculated as:

$$C_{clarity} = \frac{1}{N}\sum_{i=0}^{N} Clarity_i \tag{12}$$

where $N$ is the number of vertices on given line object.

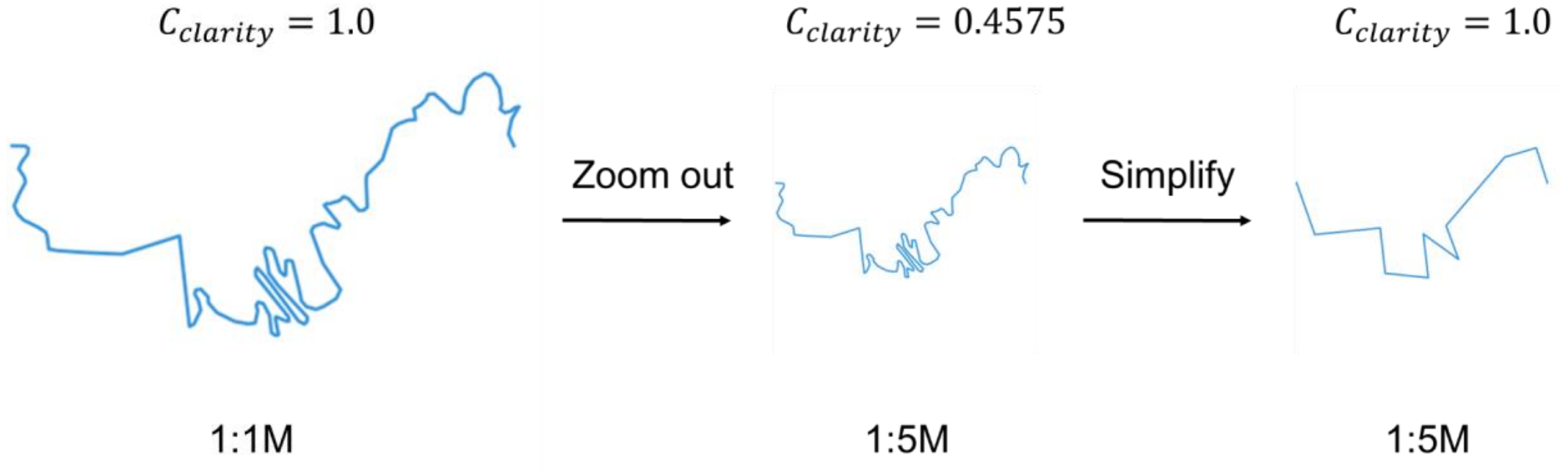

Figure 5. Example of clarity computation for line features

As shown in Figure 5, clarity decreases significantly when a line is scaled from 1:1M to 1:5M due to geometric compression, while simplification operations effectively restore clarity by removing visually redundant details. In addition to this soft evaluation, clarity may also be imposed as a hard constraint requiring all vertices of the generalized line to satisfy visibility conditions:

$$Clarity_i = \begin{cases} 0\,, & \frac{d_i^{min}}{D_{min}} < 1 \\ 1\,, & \frac{d_i^{min}}{D_{min}} \geq 1 \end{cases} \tag{13}$$

This formulation enables clarity to function either as an optimization objective or as a strict constraint within the proposed generalization control framework.

(2) Smoothness constraint: Cartographic representation integrates both scientific accuracy and visual harmony. For linear features, aesthetic quality of generalized line features is primarily reflected by geometric continuity and smoothness. Excessive angular variation may reduce visual harmony, whereas appropriate smoothing can improve cartographic appearance while maintaining essential structures.

In this study, smoothness is evaluated using vertices as computational units. As illustrated in Figure 6, the turning angle $\alpha_i$ at vertex $p_i$ is used to quantify local angular variation. Smaller turning angles indicate sharper bends and lower smoothness. The local sharpness at vertex $p_i$ is defined as

$$Sharpness_i = 0.5 \times (\cos \alpha_i + 1) \tag{14}$$

For open line features, the start and end vertices lack complete angular information and are therefore assigned a sharpness value of 0. The overall smoothness of a line feature is computed as

$$C_{smoothness} = 1 - \frac{1}{N}\sum_{i=1}^{N} Sharpness_i \tag{15}$$

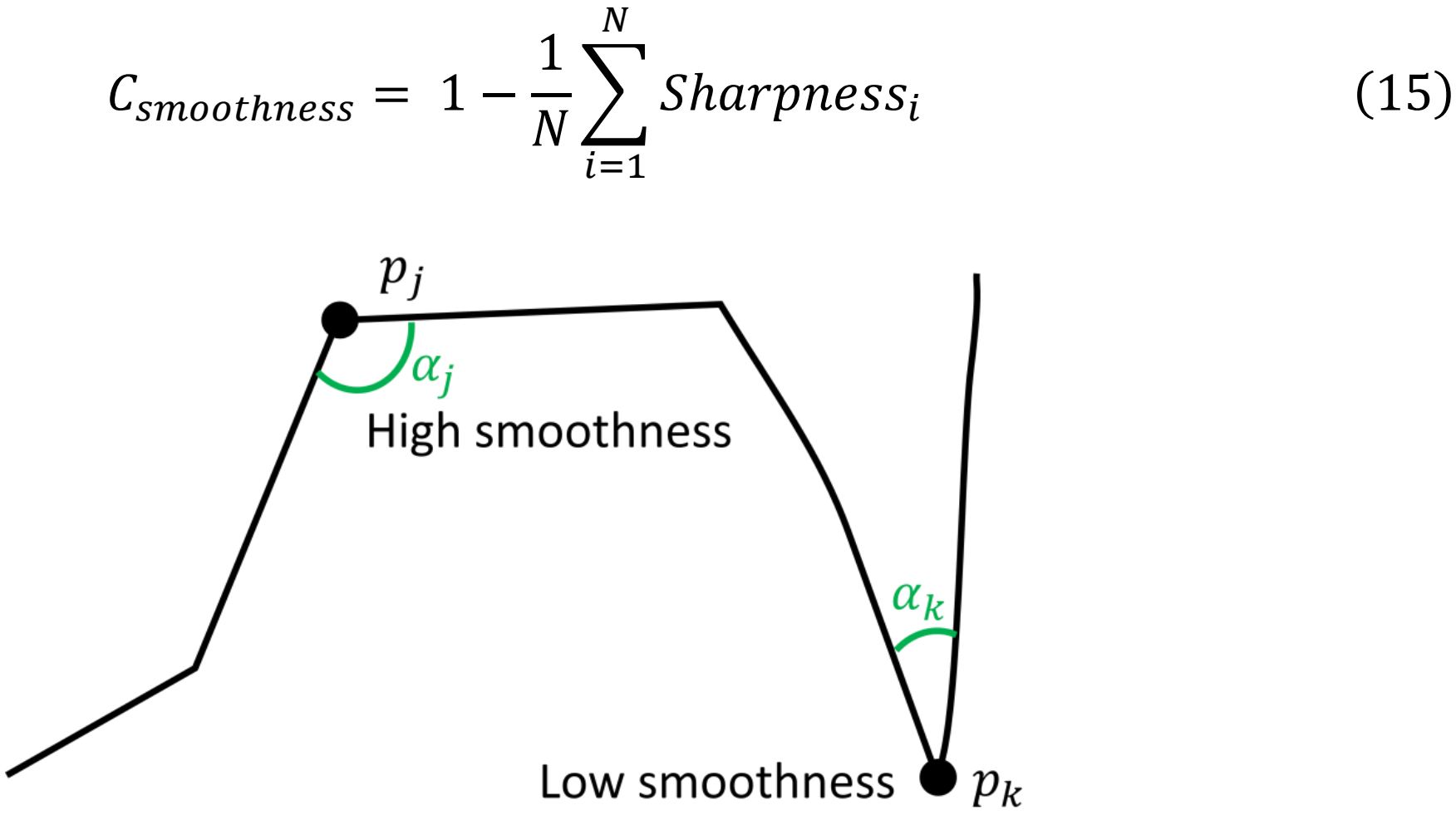

Figure 6. Schematic diagram showing the calculation for vertex smoothness

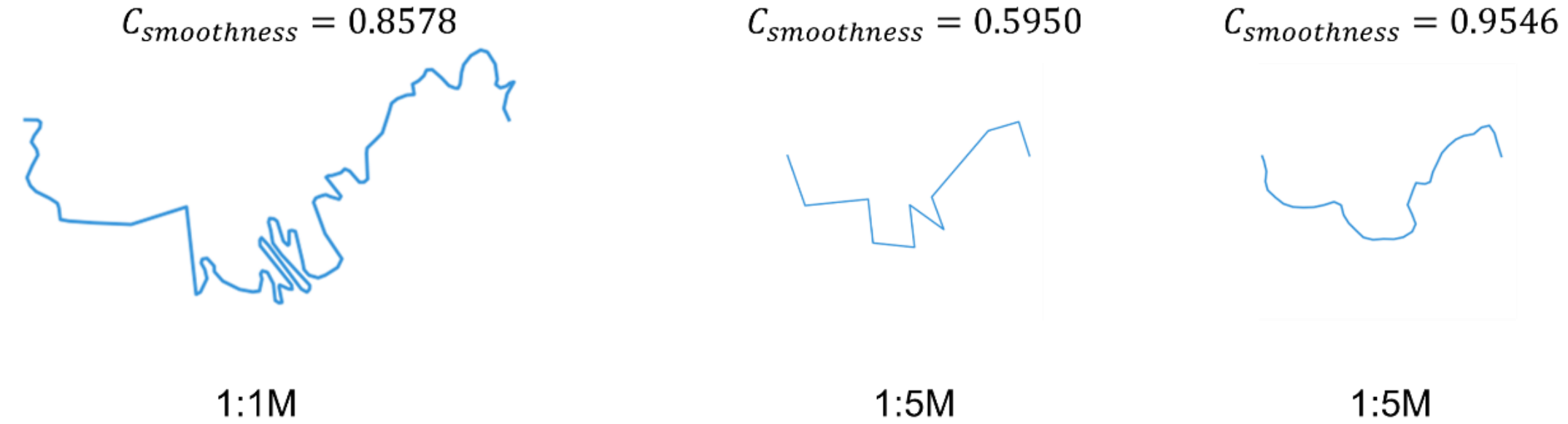

Figure 7. Example of smoothness computation for line object

As shown in Figure 7, lines simplified using the Li-Openshaw algorithm exhibit higher smoothness than those produced by the Douglas-Peucker algorithm. The computed smoothness values are consistent with visual perception, indicating that the proposed metric effectively captures aesthetic quality in line generalization.

(3) Data Correctness: Data correctness ensures the geometric and topological validity of generalized linear features. For individual line objects, a critical requirement is the absence of self-intersections (Figure 8), which represent severe topological errors that distort spatial meaning and violate cartographic consistency. Therefore, self-intersection is treated as a hard constraint that must be strictly prohibited during the generalization process. The constraint satisfaction function is defined as

$$C_{self-intersection} = \begin{cases} 0, & \text{if self} - \text{intersection} \\ 1, & \text{if no self} - \text{intersection} \end{cases} \tag{16}$$

This formulation ensures that any generalized result containing self-intersections is automatically rejected during optimization, thereby preserving data correctness.

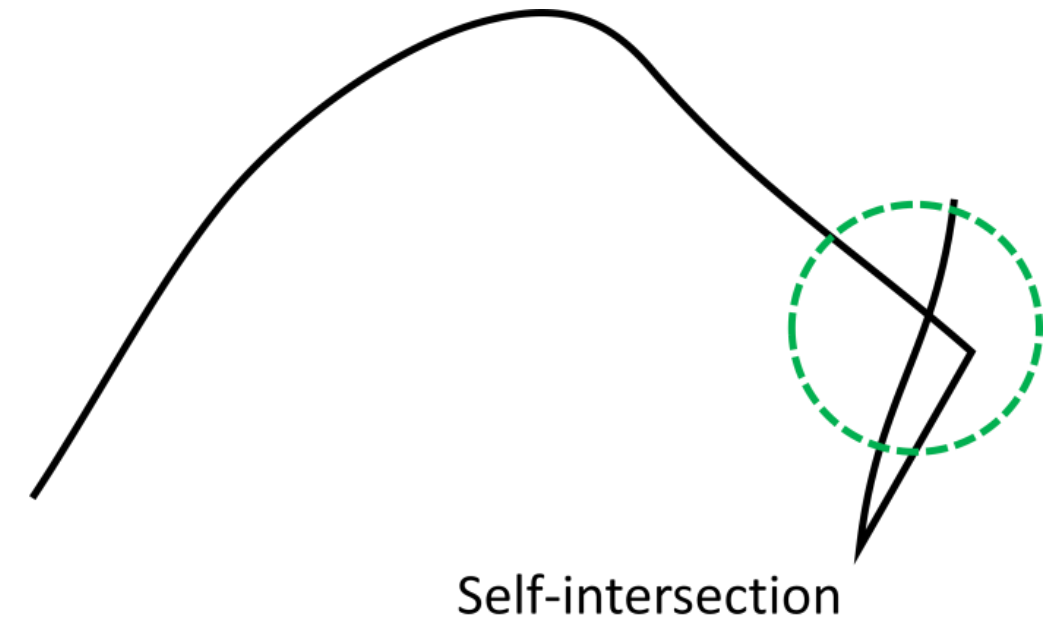

Figure 8. Example of self-intersection for line object

Within the proposed framework, cartographic generalization is formulated as a constrained optimization process balancing multiscale representation consistency and cartographic abstraction. To achieve this integration, multiscale similarity and cartographic constraints are unified into a single objective function governing adaptive parameter control during generalization.

In this study, cartographic constraints are classified into soft constraints and hard constraints, both normalized to the interval [0,1]. Soft constraints evaluate gradual satisfaction levels of cartographic requirements, including clarity, smoothness, and visual quality, whereas hard constraints represent strict feasibility conditions such as the absence of self-intersections or topological violations. Higher constraint values indicate stronger satisfaction of cartographic requirements.

When combining multiple constraints, two common aggregation strategies can be adopted: additive and multiplicative models. The weighted additive formulation,

$$C_{total} = \sum_{i=0}^{K} w_i C_i \; , \sum_{i=0}^{K} w_i = 1 \tag{17}$$

is widely used for multi-factor evaluation and $K$ is the number of constraints, $w_i$ is the corresponding weight to the constraint $C_i$. However, this approach introduces subjectivity in weight selection and cannot strictly enforce hard constraints, as violations may be compensated by other terms. To overcome these limitations, this study adopts a multiplicative aggregation model:

$$C_{total} = \prod_{i=0}^{K} C_i \tag{18}$$

The multiplicative formulation preserves normalization within the interval [0, 1]while naturally enforcing hard constraints. If any mandatory constraint is violated ($C_i = 0$), the overall constraint satisfaction becomes zero, thereby preventing infeasible solutions from being accepted during optimization. Compared with additive aggregation, the multiplicative formulation more explicitly reflects the interdependent nature of cartographic requirements during multiscale abstraction.

Multiscale similarity is subsequently integrated with constraint satisfaction using the multiplicative principle. The unified objective function is defined as：

$$T(\cdot) = Sim(l, l_g) \times C_{total}(l, l_g, S_o, S_t) \tag{19}$$

where $l$ is the original line, $l_g$ the generalized result, and $S_o$, $S_t$ denote source and target scales. Both similarity and constraint terms are normalized to [0, 1], with higher values indicating better generalization quality. For a given generalization algorithm $A(\cdot)$, optimization seeks the parameter $\boldsymbol{x}^*$ that maximizes $T(\cdot)$. In the case of line simplification, the objective function is formulated as

$$T_l(\cdot) = Sim(l, l_g) \times C_{clarity} \times C_{smoothness} \times C_{self-intersection} \tag{20}$$

Under this formulation, similarity preservation and cartographic constraints jointly regulate the scale transformation process. Multiscale similarity controls representation consistency across scales, whereas cartographic constraints regulate acceptable abstraction levels. The resulting framework therefore provides a scale-aware control for adaptive line generalization.

### ***3.5. Optimization algorithms for adaptive control***

The objective function defined in Eq.(20) is generally non-convex and algorithm-dependent because the relationship between generalization parameters, multiscale similarity, and cartographic constraints varies across different simplification algorithms and target scales. Therefore, obtaining analytical solutions through gradient-based optimization is difficult. Instead, this study formulates parameter selection as a black-box optimization problem and investigates four optimization strategies, including brute-force search, hill climbing, simulated annealing, and genetic algorithm. These methods represent different optimization paradigms, ranging from exhaustive search to local exploration and population-based global optimization.

#### *3.5.1 Brute-force search*

Brute-force search identifies the optimal parameter by enumerating candidate values from the generalization algorithm's parameter space and evaluating the corresponding objective function values. The candidate that yields the maximum objective value is selected as the optimal solution (Algorithm 1). Given sufficient sampling density over the parameter space, this method guarantees convergence to the global optimum and avoids entrapment in local minima. However, its computational cost scales rapidly with the size of the search space, making it impractical for real-world production environments. In this study, brute-force search is therefore employed primarily as a benchmark to evaluate the performance of the heuristic optimization methods.

***Algorithm 1：*** *Brute-force search*

1. Determine the maximum $p_{max}$ and minimum $p_{min}$ of the algorithm parameter space；

2. Determine the number of algorithm parameters to enumerate $m$, then the parameter control resolution $\Delta p = (p_{max} - p_{min})/(m-1)$；

3. Uniformly sample from the algorithm parameter space to generate a parameter sequence $[p_{min}, p_{min} + \Delta p, p_{min} + 2\Delta p, \cdots, p_{min} + (m-1)\Delta p]$；

4. For each parameter in the sequence, compute the objective function value to form an

objective function sequence;

5. Retrieve the maximum value from the objective function sequence, with its corresponding algorithm parameter being the optimal control parameter.

*3.5.2 Hill climbing*

Hill climbing (HC) is a greedy local search strategy that iteratively improves the current solution by exploring neighboring parameter configurations (Algorithm 2). At each iteration, it evaluates candidates in the vicinity of the current solution and moves to the one with the highest objective value. This process repeats until convergence or a predefined termination condition is met. While computationally efficient, hill climbing is highly sensitive to initialization and prone to converging to local optima, thus offering no guarantee of finding the global solution.

***Algorithm 2:*** *Hill climbing*

1. Set the maximum number of iterations to $n$;

2. Initialize the current optimal control parameter $p_{best}$ and compute its corresponding objective function value $T_{best}$;

3. Sample a candidate parameter $p_{test}$ from the neighborhood space $(p_{best} - \varepsilon, p_{best} + \varepsilon)$ of the current optimal control parameter and compute its objective function value $T_{test}$;

4. If $T_{test} > T_{best}$, update the optimal solution by setting $p_{best} = p_{test}$, $T_{best}$=$T_{test}$;

5. Terminate and output the current optimal control parameter $p_{best}$ when the iteration count reaches the predefined $n$; otherwise, return to Step 3.

*3.5.3 Simulated annealing*

Simulated annealing (SA) enhances greedy local search by incorporating a probabilistic acceptance mechanism inspired by physical annealing processes. It allows inferior solutions, those with lower objective values than the current one, to be accepted with a probability governed by a temperature parameter (Algorithm 3). This stochastic mechanism enables the algorithm to escape local optima and improves its chance of converging toward the global optimum.

***Algorithm 3:*** *Simulated annealing*

1. Set the maximum number of iterations to $n$, initial temperature to $t_{init}$, and minimum temperature to $t_{min}$;

2. Initialize the current optimal control parameter $p_{best}$ and compute its corresponding

objective function value $T_{best}$;

3. Sample a candidate parameter $p_{test}$ from the neighborhood space $(p_{best} - \varepsilon,$ $p_{best} + \varepsilon)$ of the current optimal control parameter and compute its objective function value $T_{test}$;

4. If $T_{test} > T_{best}$, update the optimal solution by setting $p_{best} = p_{test}$, $T_{best}$=$T_{test}$; if $T_{test} \leq T_{best}$, update $p_{best} = p_{test}$ with probability P, where $P = e^{\frac{(T_{test} - T_{best})}{t}} > rand()$, $t = t_{init} \times (t_{min}/t_{init})^{(k/n)}$. $k$ is the current iteration count, and $rand()$ generates a uniform random number in [0,1];

5. Terminate and output the current optimal control parameter $p_{best}$ when the iteration count reaches the predefined $n$; otherwise, return to Step 3.

### *3.5.4 Genetic algorithm*

Genetic algorithm (GA) is a population-based heuristic inspired by biological evolution. A population of candidate solutions evolves through selection, crossover, and mutation operations, with fitness evaluated directly by the objective function (Algorithm 4). GA operates on a diverse population simultaneously, which strengthens global exploration and reduces susceptibility to local optima.

***Algorithm 4:*** *Genetic algorithm*

1. Set the evolutionary generation count to $n$, population size to $m$, crossover rate to $R_{crossover}$, and mutation rate to $R_{mutation}$;

2. Initialize the current parameter population $\{p_1, p_2, \cdots, p_m\}$, evaluate their corresponding objective function values $\{T_1, T_2, \cdots, T_m\}$, and set the initial optimal control parameter $p_{best}$ as the individual with maximal objective value, where $T_{best} = max\{T_1, T_2, \cdots, T_m\}$.

3. Generate the next-generation population by fitness-proportional selection, where the selection probability of each individual is weighted by its objective function value;

4. With a probability of $R_{crossover}$, crossover is performed on individuals in the next generation. The crossover method is averaging, i.e., $p_{new} = 0.5 \times (p_i + p_j)$;

5. With a probability of $R_{mutation}$, mutation is applied to individuals in the next generation. The mutation method involves random sampling within the parameter space;

6. Calculate the objective function value for each parameter in the next generation. If the maximum value of the objective function exceeds $T_{best}$, the corresponding parameter is set as the current optimal control parameter $p_{best}$;

7. When the number of generations reaches the predefined value $n$ output the current optimal control parameter $p_{best}$; otherwise, return to Step 3.

## 4. Experiments

### *4.1. Dataset and method configurations*

The proposed framework is evaluated using line simplification as a representative application in cartographic generalization. A manually curated dataset comprising 100 diverse line features at an original scale of 1:1M was simplified to six target scales: 1:2.5M, 1:5M, 1:10M, 1:25M, 1:50M, and 1:100M. The dataset covers multiple geographic feature types, including rivers, roads, and contours, ensuring a broad representation of geometric characteristics. Figure 9 presents these 100 original line features, showcasing their varied geometric patterns and complexity.

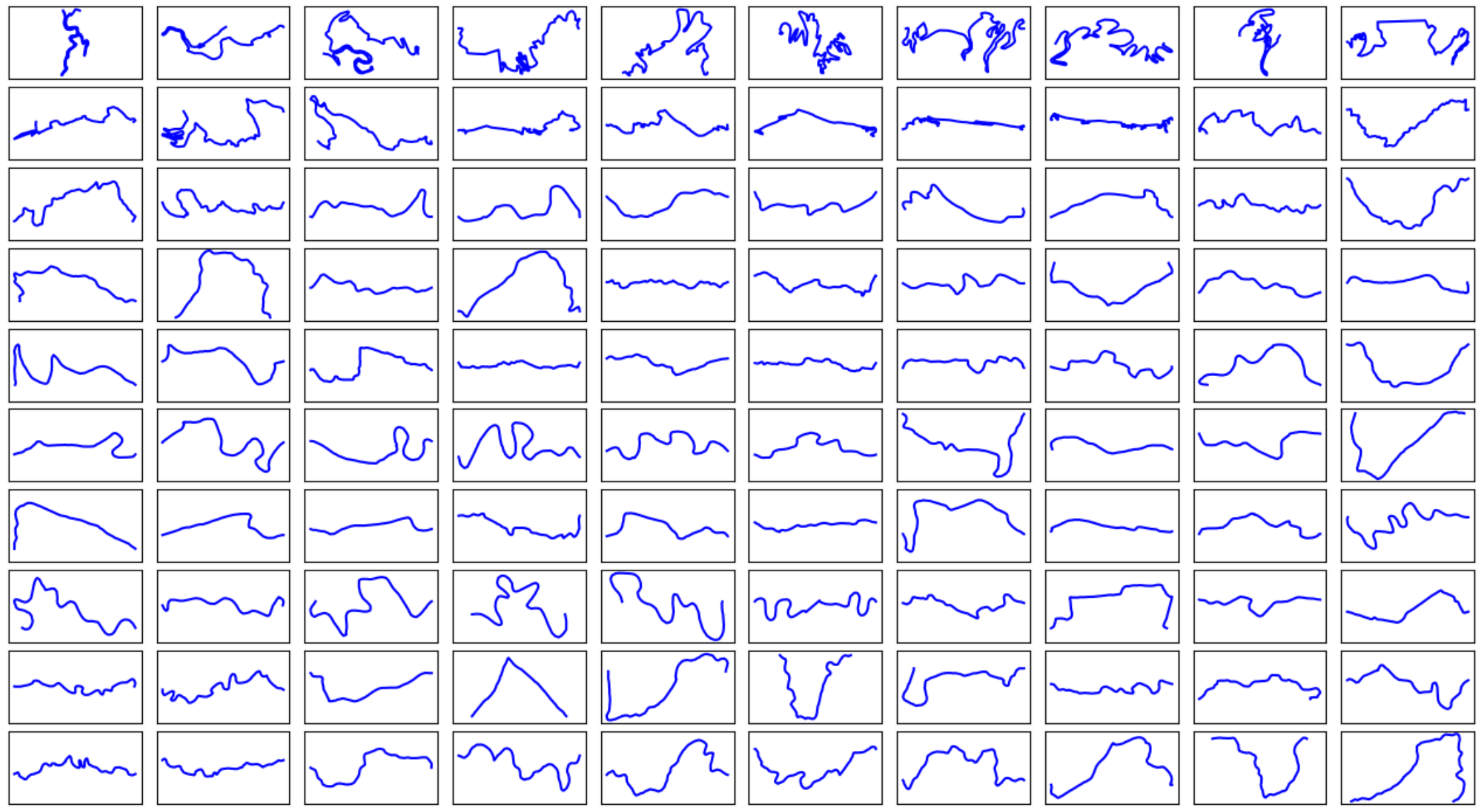

Figure 9. The 100 diverse original line features used in the evaluation.

Three classical line simplification algorithms: Douglas-Peucker (DP), Visvalingam-Whyatt (VW), and Li-Openshaw, are employed to evaluate the framework's adaptability, with their parameters automatically controlled under the proposed optimization objective. Information preservation is quantified via the DML-based line similarity measure (Li et al., 2023), supplemented by two geometric similarity metrics: the median Hausdorff distance and the Fréchet distance. Constraint satisfaction is measured according to the composite model described in Section 3.4, covering clarity, smoothness, and self-intersection. Unless otherwise stated, all control curves in Figure 10 follow the unified color scheme illustrated in Figure 9, where each curve denotes a distinct evaluation metric.

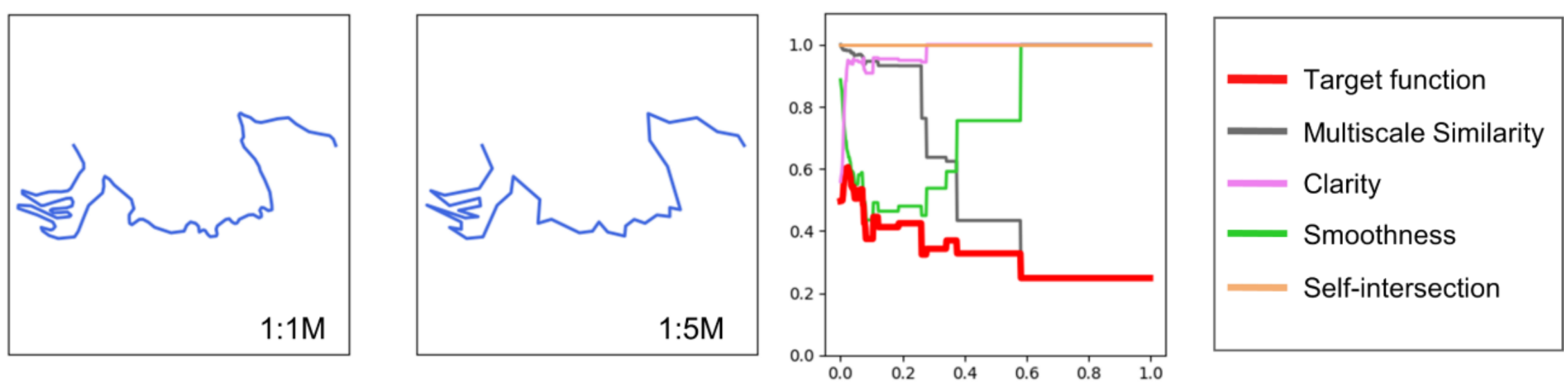

Figure 10. Legend and metric definitions for line simplification control curves

### *4.2. Process control analysis of DP algorithms*

This experiment employs a brute-force search strategy, which enables a comprehensive visualization of the control mechanism of the Douglas-Peucker (DP) algorithm. The variation curves of the objective function, similarity, clarity, smoothness, and self-intersection constraints are examined across multiple target scales. Figure 11 presents a representative example, while Table 1 summarizes the corresponding control metric statistics.

As illustrated in Figure 11, the framework consistently identifies the optimal simplification state for the DP algorithm at each target scale, characterized by the maximum value of the objective function. The optimal tolerance threshold increases as the target scale becomes smaller, indicating that stronger simplification is required at reduced scales. Concurrently, the similarity values at the optimal control state exhibit a gradual decline, reflecting the inherent balance between preserving geometric information and accommodating scale reduction during multi-scale transformation.

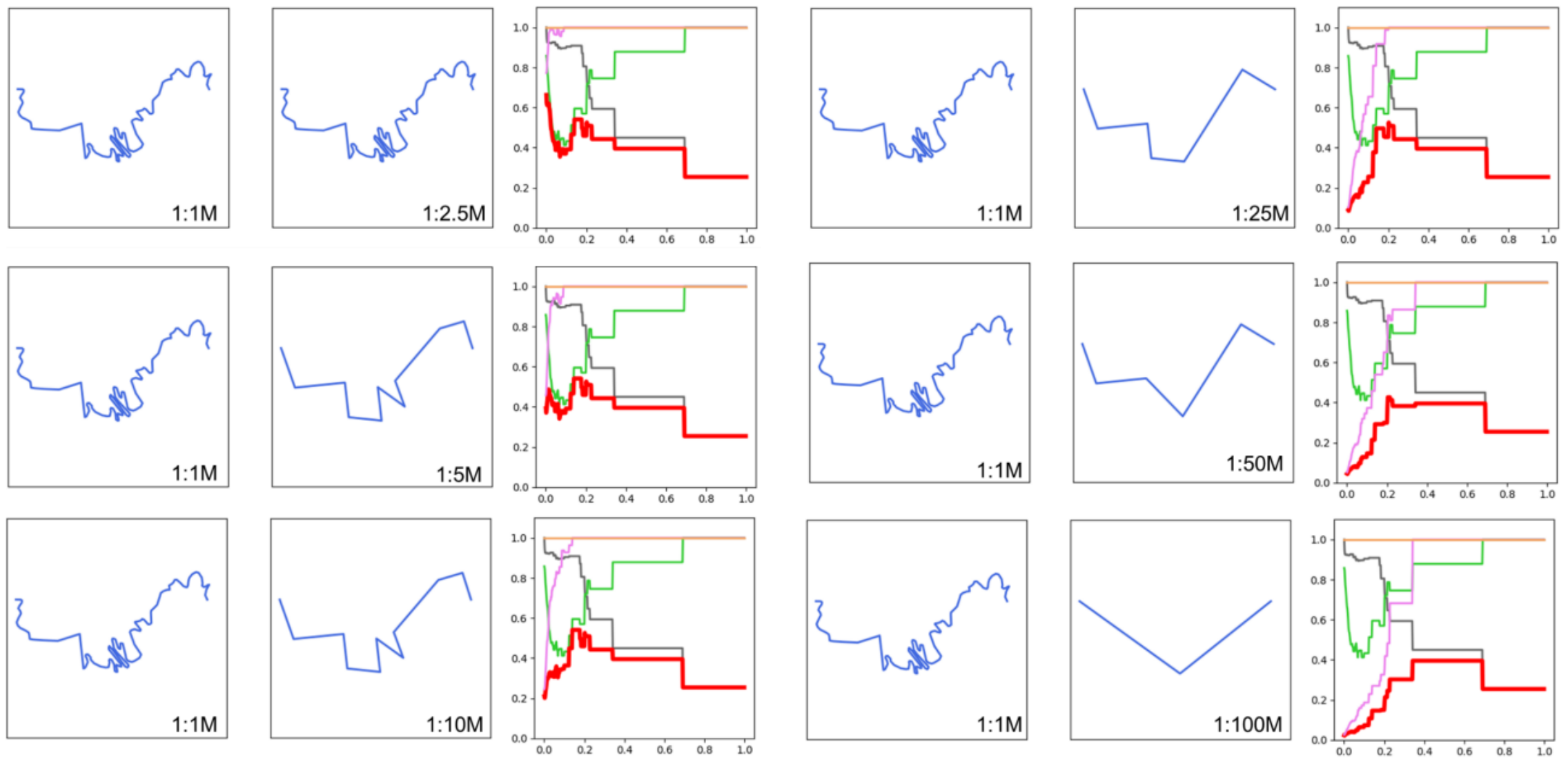

Figure 11. Process control examples of the DP algorithm at different target map scales

Table 1. Statistics of DP algorithm process control cases at different target scales

| **Target scale** | ***T*max** | **Similarity** | **Initial Clarity** | **Final Clarity** |
|---|---|---|---|---|
| 1:2.5M | **0.6631** | 1.0 | 0.7730 | 0.7730 |
| 1:5M | **0.5407** | 0.9088 | 0.4575 | 1.0 |
| 1:10M | **0.5407** | 0.9088 | 0.2443 | 1.0 |
| 1:25M | **0.5258** | 0.7414 | 0.1025 | 1.0 |
| 1:50M | **0.4259** | 0.7081 | 0.0440 | 0.8348 |
| 1:100M | **0.3949** | 0.4494 | 0.0256 | 1.0 |

The clarity function plays a dominant role in linking algorithm parameters to map scale. As the target scale decreases, the initial clarity value progressively declines, and the peak of the clarity curve shifts toward larger simplification thresholds. Because clarity is implemented as a soft constraint, optimal solutions at scales of 1:2.5M and 1:50M do not always achieve full legibility (clarity < 1), demonstrating the framework's ability to balance competing objectives.

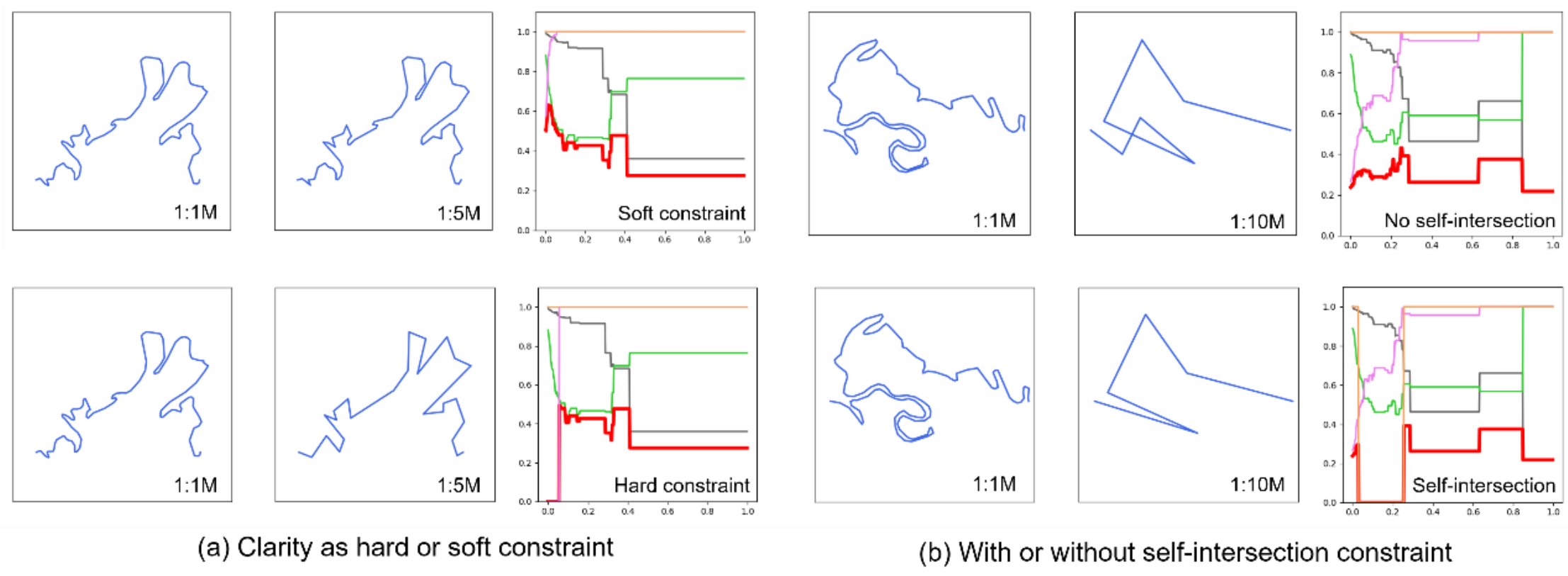

Figure 12. Comparison between varying constraint configurations

Figure 12(a) compares results obtained under soft and hard clarity constraints. When clarity is treated as a hard constraint, the control process enforces stronger simplification to guarantee legibility at the target scale. Figure 12(b) further illustrates the role of the self-intersection constraint. Without this constraint, certain parameter ranges produce invalid geometries containing self-intersections; incorporating the hard constraint effectively restricts the optimization space and prevents topological errors. These results demonstrate that different constraint configurations can flexibly regulate algorithm behavior according to application requirements.

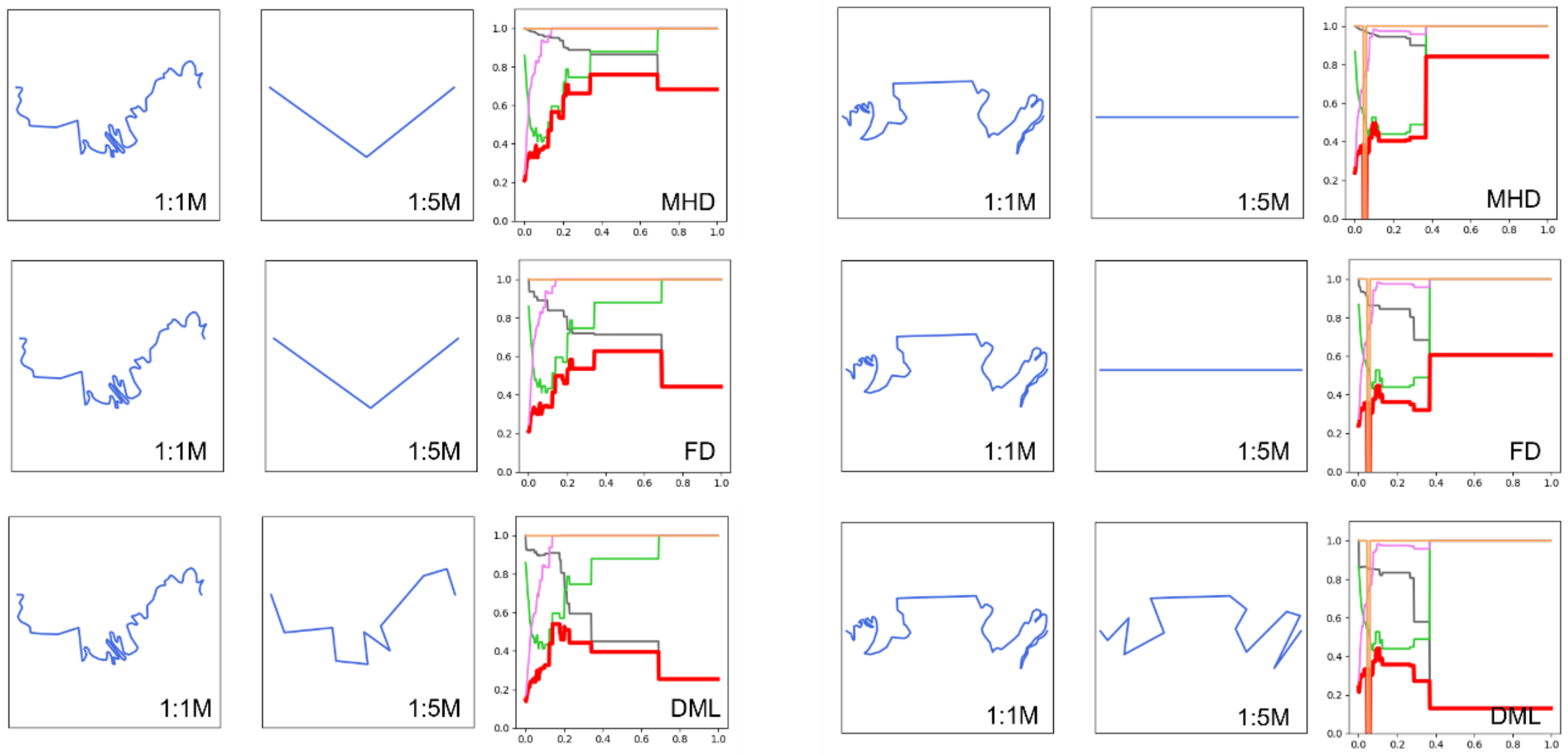

Figure 13. Control results under different similarity metrics at the target scale of 1:5M

Figure 13 evaluates the influence of different similarity measures. The similarity metric derived from deep metric learning shows greater sensitivity to structural changes than median Hausdorff distance and Fréchet distance. The resulting similarity curve responds more sharply to significant shape variations, aligning better with human perceptual judgment and producing more balanced control outcomes. Across diverse line features, the automatically determined simplification parameters consistently produce reasonable generalization results, demonstrating the effectiveness of the similarity-driven optimization framework for automated control of the DP algorithm.

### *4.3. Comparison between multiple simplification algorithms*

This experiment compares the control performance of three classical line simplification algorithms: Douglas-Peucker, Visvalingam-Whyatt, and Li-Openshaw. Figure 14 presents representative simplification results under the proposed control framework, and Table 2 summarizes the corresponding maximum objective function values and key metrics.

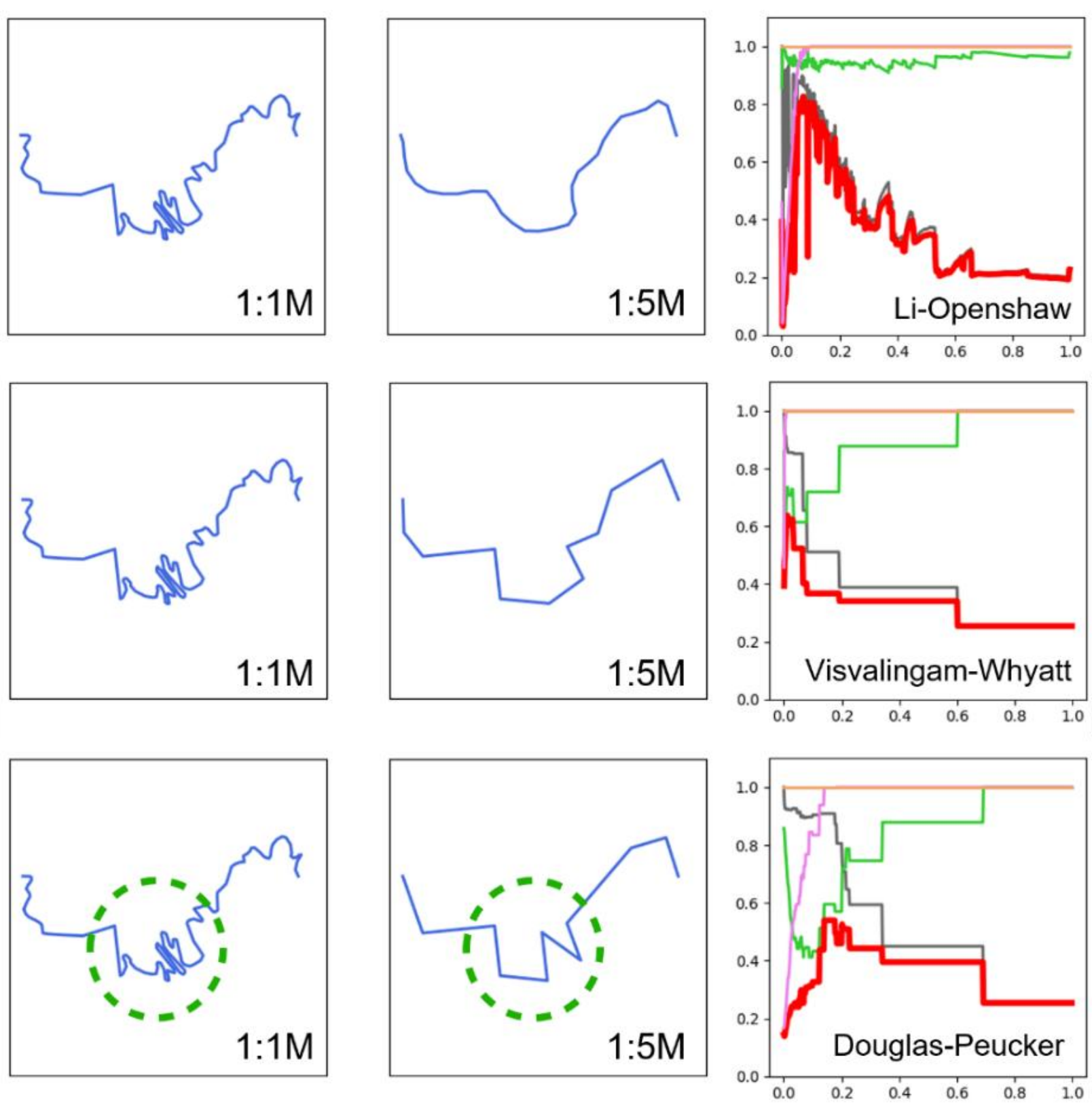

Figure 14. Control results and process curves of different simplification algorithms

Table 2. Results for different algorithms at the target scale of 1:5M

| **Algorithms** | ***T*max** | **Similarity** | **Initial Clarity** | **Final Clarity** | **Smoothness** |
|---|---|---|---|---|---|
| Li-Openshaw | 0.8261 | 0.8934 | 0.4575 | 0.9892 | 0.9348 |
| DP | 0.5407 | 0.9088 | 0.4575 | 1.0 | 0.5950 |
| VW | 0.6365 | 0.8652 | 0.4575 | 1.0 | 0.7357 |

Overall, the Li-Openshaw algorithm achieves the highest objective function values (*Tmax*) across most target scales, followed by the VW algorithm, while the DP algorithm shows comparatively lower overall performance. This ranking is consistent with visual inspection. Both Li-Openshaw and VW generate smoother generalized lines while maintaining global shape consistency, which is reflected in their higher smoothness scores. In particular, Li-Openshaw, as a scale-driven smoothing method, consistently maintains high smoothness values across scales, contributing to its superior overall objective performance.

Interestingly, the similarity scores reveal a different pattern: the DP algorithm achieves relatively higher similarity values, whereas Li-Openshaw yields lower similarity scores. This difference arises because DP preserve original and critical vertices during simplification, retaining more local geometric details. As highlighted in the circled regions of Figure 14, the DP algorithm better preserves curvature and directional changes, resulting in higher structural similarity despite lower smoothness.

By integrating similarity and constraint satisfaction within a unified framework, the method enables systematic evaluation and adaptive control across different generalization algorithms, rather than favoring a single metric. The experiment confirms the flexibility and general applicability of the proposed control framework for multi-algorithm process regulation.

### *4.4. Robustness under different optimization strategies*

To evaluate the robustness of the proposed similarity-driven control framework, three representative optimization strategies, including Hill Climbing (HC), Simulated Annealing (SA), and Genetic Algorithm (GA), were employed for parameter search. The purpose of this comparison is not to benchmark optimization algorithms themselves, but to examine whether the proposed framework can consistently identify balanced generalization solutions under different optimization conditions.

A brute-force search strategy was first employed to establish a reference benchmark for optimal control. Specifically, 500 uniformly sampled parameter values were enumerated at each target scale, and the parameter configuration achieving the maximum objective function value was regarded as the reference optimal solution. The HC, SA, and GA methods were then evaluated according to their ability to approach this benchmark for the DP and VW algorithms across 100 line objects at six target scales. Experiments were conducted under 50 and 100 iterations using multiple control precision thresholds ranging from ±0.01 to ±0.2.

Results presented in Tables 3–5 show that increasing iteration numbers consistently improves optimization success rates for all optimization strategies under different precision thresholds. This observation indicates that the proposed similarity-driven objective function provides a stable optimization landscape for parameter search across multiple scales. Although all three optimization strategies are capable of approaching the reference benchmark, their convergence behaviors differ substantially.

Table 3. Reach rate for DP algorithm across 100 simplification cases（Initial value is $D_{min}$）

| **Optimization Method（Iterative Number）** | **Reach Rate ($\pm 0.01$)** | **Reach Rate ($\pm 0.05$)** | **Reach Rate ($\pm 0.1$)** | **Reach Rate ($\pm 0.2$)** |
|---|---|---|---|---|
| **HC（n=50）** | 0.662 | 0.872 | 0.928 | 0.970 |
| **HC（n=100）** | 0.760 | 0.896 | 0.940 | 0.970 |
| **SA（n=50）** | 0.610 | 0.886 | 0.940 | 0.982 |
| **SA（n=100）** | 0.700 | 0.932 | 0.966 | 0.994 |
| **GA（n=50）** | 0.838 | 0.958 | 0.990 | 0.994 |
| **GA（n=100）** | **0.916** | **0.970** | **0.990** | **0.998** |

Table 4. Reach rate for VW algorithm across 100 simplification cases（Initial value is $D_{min}$）

| **Optimization Method （Iterative Number）** | **Reach Rate ($\pm 0.01$)** | **Reach Rate ($\pm 0.05$)** | **Reach Rate ($\pm 0.1$)** | **Reach Rate ($\pm 0.2$)** |
|---|---|---|---|---|
| **HC（n=50）** | 0.406 | 0.666 | 0.816 | 0.930 |
| **HC（n=100）** | 0.568 | 0.800 | 0.876 | 0.944 |
| **SA（n=50）** | 0.360 | 0.606 | 0.860 | 0.940 |
| **SA（n=100）** | 0.450 | 0.712 | 0.884 | 0.964 |
| **GA（n=50）** | 0.478 | 0.810 | 0.942 | 0.990 |
| **GA（n=100）** | **0.582** | **0.896** | **0.978** | **0.998** |

Table 5. Reach rate for VW algorithm across 100 simplification cases（Initial value is $0.5D_{min}^2$）

| **Optimization Method （Iterative Number）** | **Reach Rate ($\pm 0.01$)** | **Reach Rate ($\pm 0.05$)** | **Reach Rate ($\pm 0.1$)** | **Reach Rate ($\pm 0.2$)** |
|---|---|---|---|---|
| **HC（n=50）** | 0.560 | 0.860 | 0.958 | 0.996 |
| **HC（n=100）** | 0.668 | 0.926 | 0.980 | 0.996 |
| **SA（n=50）** | 0.616 | 0.868 | 0.952 | 0.990 |
| **SA（n=100）** | 0.663 | 0.892 | 0.966 | 0.998 |
| **GA（n=50）** | 0.854 | 0.988 | 1.0 | 1.0 |
| **GA（n=100）** | **0.890** | **0.998** | **1.0** | **1.0** |

Among the evaluated methods, GA achieves the highest success rates across nearly all precision thresholds and iteration settings, demonstrating stronger robustness in identifying balanced parameter configurations for multiscale generalization control. HC performs competitively under strict precision thresholds when iteration numbers are limited, but it frequently converges to local optima as the complexity of parameter search increases. SA demonstrates improved global exploration capability compared with HC; however, under limited iteration conditions, its optimization advantage remains less stable. Figure 15 further illustrates that GA more consistently approaches near-optimal control states with fewer iterations, whereas HC and SA generally require additional iterations to achieve comparable optimization performance.

The experiments also reveal that optimization initialization significantly affects framework performance, particularly for the VW algorithm. As shown in Table 5, using $0.5D_{min}^2$ as the initialization condition substantially improves both optimization precision and success rates compared with using $D_{min}$ directly. This observation suggests that initialization sensitivity remains an important factor influencing optimization efficiency in similarity-driven generalization control.

Overall, the experimental results demonstrate that the proposed framework maintains stable and interpretable control behavior under different optimization mechanisms. Although the optimization algorithms differ in convergence efficiency and search stability, the resulting generalized representations consistently preserve the balance between multiscale representation consistency and cartographic abstraction. These results suggest that the effectiveness of the proposed approach primarily derives from the unified similarity–constraint formulation rather than from any specific optimization strategy.

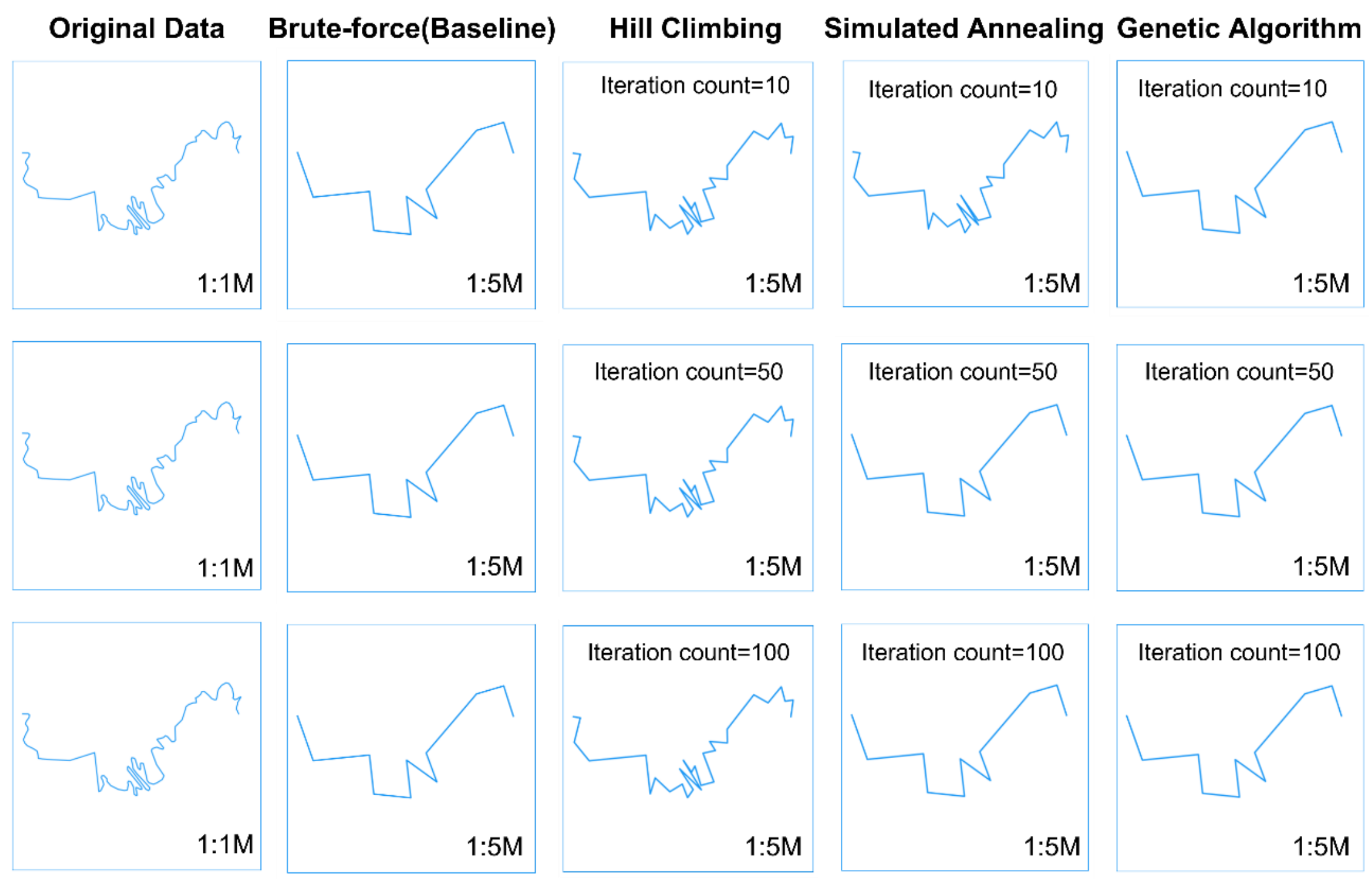

Figure 15. Control result examples of the DP algorithm using different optimization methods

### *4.5. Adaptive simplification on line groups*

The proposed generalization framework further supports adaptive simplification of line groups. Through a two-level optimization strategy: parameter selection and algorithm combination, it aims at achieving globally optimal results. At the parameter level, the framework abandons fixed thresholds or simple statistical values; instead, it dynamically determines the simplification tolerance for each individual line in the group based on local geometric characteristics, thus striking a better balance between preserving key shape features and enhancing visual clarity. At the algorithm level, the framework does not rely on a single simplification method; rather, it adaptively selects the most suitable algorithm, among DP, VW, and Li-Openshaw in our case, according to the best target value of each line. By this dual-layer adaptive mechanism, the framework automatically trades off similarity, visual clarity, smoothness, and topological consistency at the target scale, ultimately achieving the best overall simplification quality. The following two experiments validate the effectiveness of parameter-level and algorithm-level adaptivity, respectively.

Table 6. Average performance metrics of DP simplification with different thresholds

| Methods | *T*max | Similarity | Clarity | Smoothness | Self-Intersection |
|---|---|---|---|---|---|
| **Group1(Adaptive)** | **0.6966** | **0.7950** | **0.9411** | **0.9403** | **1.0** |
| Group1(Median) | 0.5407 | 0.7178 | 0.8679 | 0.9285 | 1.0 |
| Group1(Mean) | 0.6365 | 0.6382 | 0.9006 | 0.9273 | 1.0 |
| **Group2(Adaptive)** | **0.6718** | **0.7278** | **0.9615** | **0.9598** | **1.0** |
| Group2(Median) | 0.5407 | 0.7060 | 0.8138 | 0.9127 | 1.0 |
| Group2(Mean) | 0.5396 | 0.6537 | 0.8776 | 0.9295 | 1.0 |

*4.5.1 Adaptive Threshold Determination in DP Simplification*

Table 6 presents the results of applying the DP algorithm with three different tolerance thresholds, including adaptive, median, and mean, to two line groups at the target scale of 1:5M. Group 1 belongs to a river basin, while Group 2 is part of a road network. Overall, the adaptive threshold consistently outperforms the fixed statistical thresholds (median and mean) across most evaluation metrics. For Group 1, the adaptive method achieves the highest $Tmax$ (0.6966), Similarity (0.7950), and Clarity (0.9411), while maintaining the same perfect Smoothness (0.9403) and Self-Intersection avoidance (1.0) as the median method. As shown in Figure 16, the optimal epsilon distribution confirms the adaptive nature of the tolerance: it effectively overcomes the limitations of fixed thresholds, where a small tolerance leads to under-simplification (retaining excessive detail and noise) and large tolerance results in over-simplification (losing critical shape characteristics and distorting the original geometry).

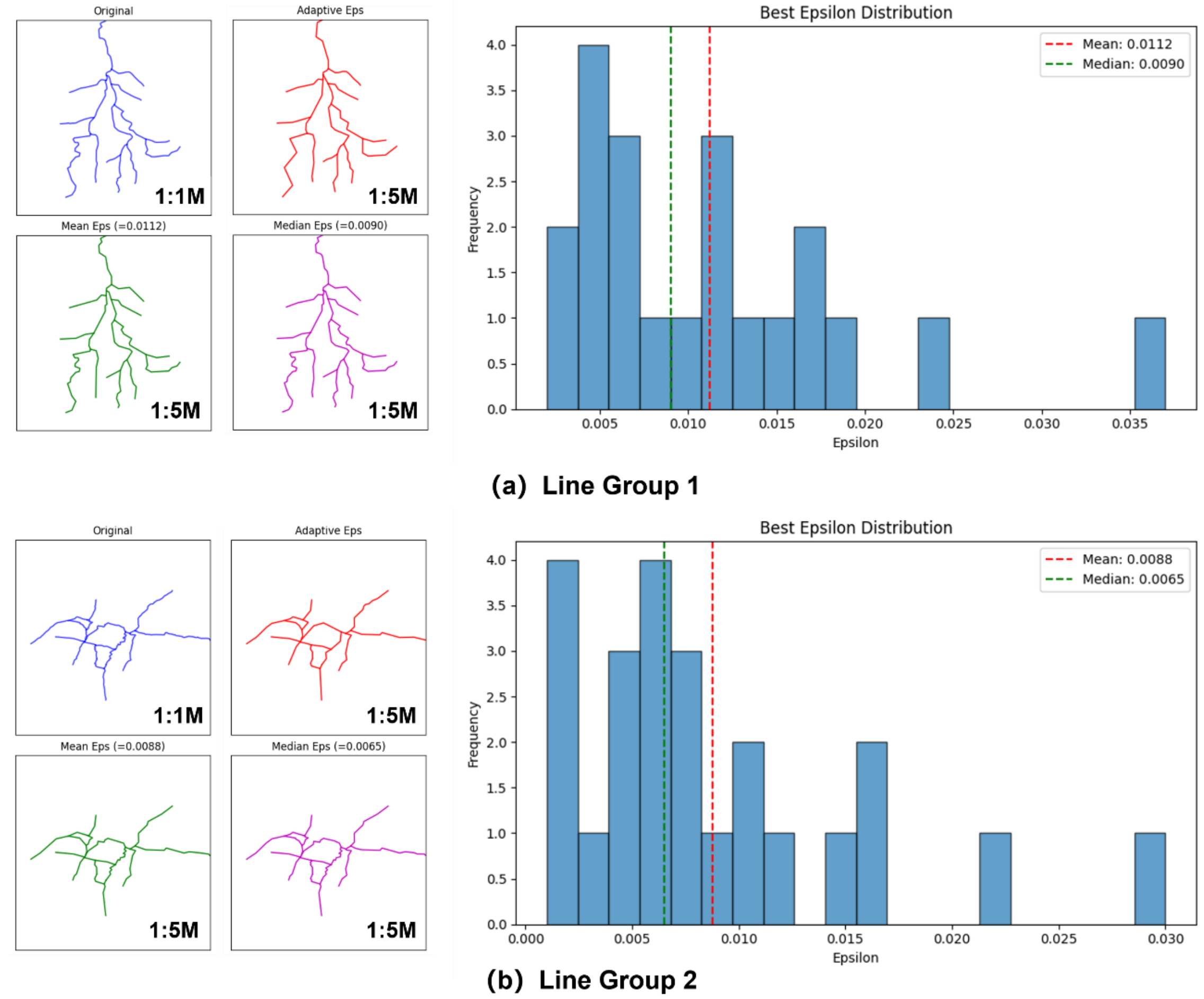

Figure 16. Illustrations of adaptive DP simplification on line groups

For Group 2, the superiority of the adaptive approach is even more pronounced. It records the highest Tmax (0.6718), Similarity (0.7278), Clarity (0.9615), and Smoothness (0.9598). In contrast, both median and mean methods produce markedly lower similarity scores (0.7060 and 0.6537, respectively) and lower clarity values, with the mean method showing the lowest Tmax (0.5396). These results collectively demonstrate that the adaptive threshold, which dynamically considers the local geometric characteristics of each line, effectively balances positional accuracy (Similarity) and visual quality (Clarity and Smoothness) far better than a one-parameter-fits-all statistical approach.

Table 7. Average performance metrics of varying algorithms with adaptive thresholds

| **Algorithms** | ***T*max** | **Similarity** | **Clarity** | **Smoothness** | **Self-Intersection** |
|---|---|---|---|---|---|
| Group1(Hybrid) | **0.7228** | **0.8153** | 0.9360 | 0.9520 | 1.0 |
| Group1(DP) | 0.6966 | 0.7950 | 0.9411 | 0.9403 | 1.0 |
| Group1(VW) | 0.6468 | 0.6963 | **0.9929** | 0.9427 | 1.0 |
| Group1(Li-Openshaw) | 0.6413 | 0.6982 | 0.9308 | **0.9855** | 1.0 |
| Group2(Hybrid) | **0.6882** | 0.7263 | 0.9728 | 0.9750 | 1.0 |
| Group2(DP) | 0.6718 | **0.7278** | 0.9615 | 0.9598 | 1.0 |
| Group2(VW) | 0.6078 | 0.6288 | **1.0000** | 0.9700 | 1.0 |
| Group2(Li-Openshaw) | 0.6271 | 0.6435 | 0.9793 | **0.9932** | 1.0 |

#### *4.5.2 Adaptive Algorithm Combination with Adaptive Parameters*

Table 7 compares the hybrid adaptive algorithm, which dynamically selects among DP, VW, and Li-Openshaw based on line group properties, with each algorithm operating under its own optimally adapted tolerance, against its three individual algorithms. Figure 16 presents the corresponding visual simplification results produced by the four algorithms at the target scale of 1:5M, providing a qualitative complement to the quantitative metrics reported in Table 7. The most prominent advantage of the hybrid method lies in its superior Tmax values, which represent the overall objective function that comprehensively balances similarity, clarity, smoothness, and topological constraints.

For both test groups, the hybrid algorithm consistently achieves the highest *Tmax*, 0.7228 for Group 1 and 0.6882 for Group 2, substantially outperforming DP (0.6966 and 0.6718), VW (0.6468 and 0.6078), and Li-Openshaw (0.6413 and 0.6271). This consistent advantage stems from the hybrid's ability to reconcile the trade-off among similarity, clarity, and smoothness, attaining the best composite score in both cases. This quantitative trend is visually corroborated in Figure 16. The hybrid output exhibits the most faithful representation of the original line groups, preserving critical bends in the river network (Group 1) while retaining the essential structure of road networks (Group 2) without introducing jagged artifacts or excessive rounding. In contrast, the individual methods each exhibit deficiencies: DP retains excessive angular artifacts; VW tends to oversimplify geometrically significant segments; and Li-Openshaw produces overly rounded shapes that deviate considerably from the original geometry. These visual comparisons further confirm that the hybrid method achieves a more desirable balance than any single algorithm, demonstrating its robust adaptability across different line group types.

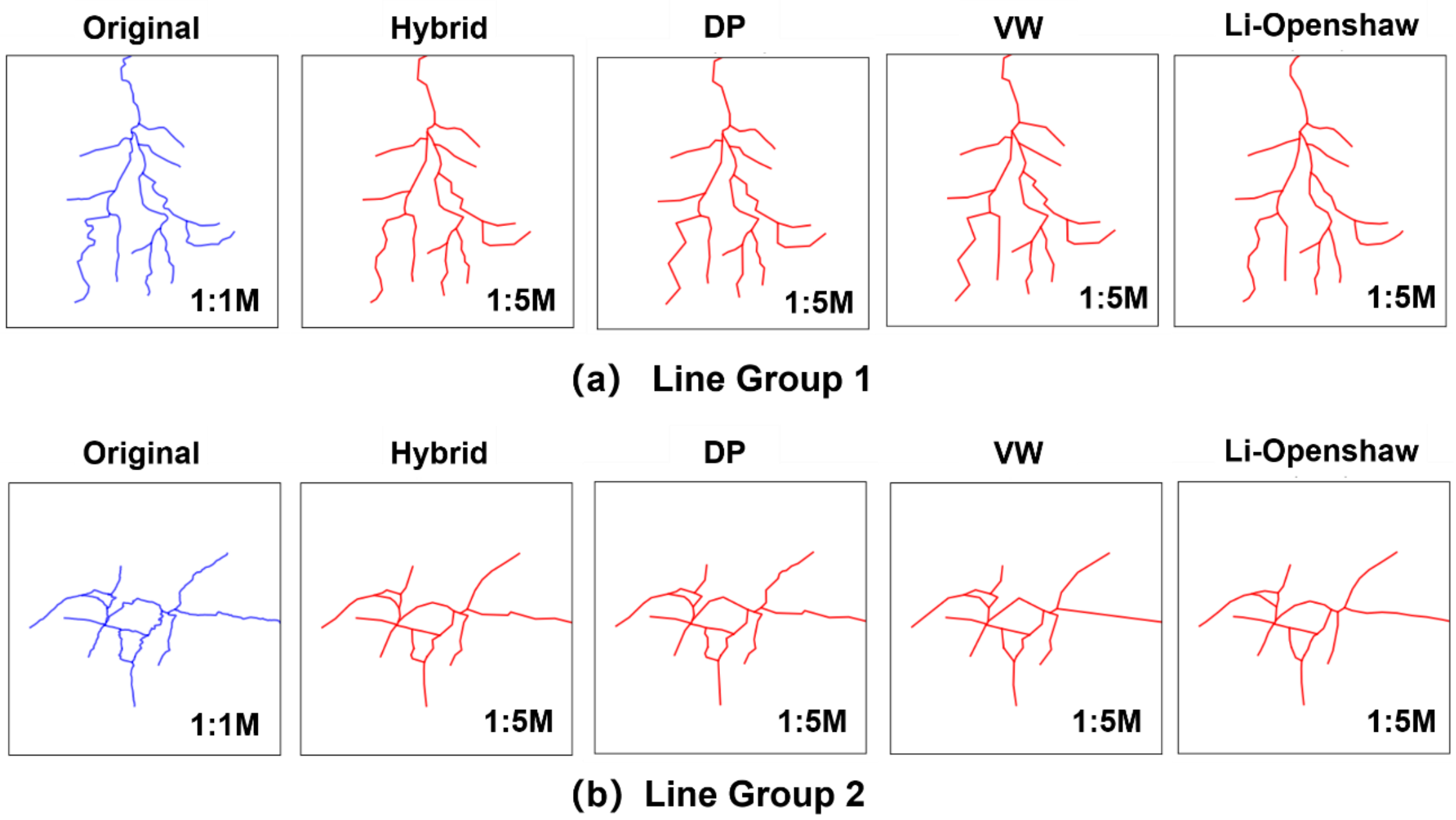

Figure 17. Simplification results with different algorithms at the target scale of 1:5M

Table 7 indicates that each individual algorithm exhibits a distinct strength profile: DP preserves Similarity, VW maximizes Clarity, and Li-Openshaw optimizes Smoothness, as clearly observed in Figure 17. The heterogeneous composition of each line group further demonstrates the framework's ability to assign the most suitable method per line, thereby effectively harnessing complementary strengths to maximize the overall objective value.

The entire pipeline from parameter adaptation to algorithm selection is fully automatic within our framework. The optimization method systematically explores the parameter space and algorithm combinations without manual intervention, effectively simulating the iterative trial-and-error process of human cartographers. By replicating this expert workflow in a data-driven manner, the framework eliminates subjective bias and substantially reduces manual tuning effort. This confirms that our approach not only solves the multi-objective optimization problem, but also offers a practical and scalable solution for cartographic production.

## 5. Discussion

### *5.1. Implications of balancing multiscale similarity and cartographic constraints*

Cartographic generalization has traditionally been investigated from three relatively independent perspectives: preserving similarity between original and generalized representations, satisfying cartographic constraints, and designing algorithms for geometric transformation. This study provides an integrated perspective by formulating generalization control as a constrained similarity-driven optimization problem, where multiscale similarity and cartographic requirements jointly determine the optimal transformation process.

The experimental results demonstrate that similarity preservation alone is insufficient for controlling generalization behavior. Although higher similarity indicates stronger retention of original spatial characteristics, excessive similarity may restrict necessary abstraction and reduce cartographic readability at smaller scales. Conversely, cartographic constraints alone cannot explicitly quantify information loss during scale transformation. By integrating similarity and constraints within a unified objective function, the proposed framework establishes a quantitative mechanism to balance information preservation and cartographic abstraction. From this perspective, the methodological implications of the framework can be summarized as follows.

(1) A unified control mechanism independent of specific generalization algorithms.

Rather than optimizing individual algorithms separately, the proposed framework abstracts generalization as a parameter optimization problem under a common objective function. The optimization process is therefore decoupled from specific simplification strategies, enabling different algorithms, such as Douglas-Peucker, Visvalingam-Whyatt, and Li-Openshaw, to be evaluated and controlled within the same framework. This provides a potential pathway toward more scalable and interoperable automated generalization systems.

(2) A quantitative connection between similarity assessment and process control.

Previous studies have mainly employed spatial similarity as an evaluation criterion after generalization. This study extends its role by introducing similarity as an active optimization objective that guides parameter selection. Such a formulation establishes a direct link between the assessment of generalized representations and the control of generalization processes, transforming similarity from a passive measurement into a decision-making mechanism.

(3) A scale-aware optimization strategy for adaptive generalization.

The proposed framework enables automatic identification of parameter configurations under different target scales by balancing similarity preservation and cartographic constraints. Compared with manual parameter tuning, which requires repeated expert intervention, the optimization-based approach provides a more systematic and interpretable strategy for adaptive generalization. This perspective supports the development of autonomous mapping workflows in which generalization decisions can be optimized according to target-scale requirements.

Overall, the contribution of this study is not to replace existing generalization algorithms, but to provide an additional control layer that integrates evaluation, optimization, and algorithm execution within a unified framework.

### *5.2. Limitations and future research directions*

Despite the encouraging results, several limitations remain and point to important directions for future research. These limitations are addressed below in terms of generalization scope, operator generality, constraint formulation, and optimization efficiency.

First, the current framework focuses primarily on single-feature generalization, where similarity and constraints are evaluated at the level of individual line objects. However, real-world map generalization often involves interactions among multiple geographic features, including roads, rivers, boundaries, and urban objects. Future research should extend the framework toward group-level and multi-feature generalization by incorporating relational constraints, such as proximity, alignment, connectivity, and spatial pattern preservation. In addition, similarity measures should be further developed to capture collective structural consistency rather than only individual feature similarity.

Second, although the proposed framework is designed to be algorithm-independent, current validation is mainly conducted on line simplification algorithms. The applicability of the framework to other generalization operators, including displacement, aggregation, selection, and typification, requires further investigation. Such extensions will help evaluate whether similarity-driven optimization can serve as a general control paradigm across different feature types and transformation processes.

Third, the current framework relies on manually designed cartographic constraints, including clarity, smoothness, and topological validity. Although these constraints effectively represent important cartographic requirements, they may not fully capture complex perceptual judgments made by expert cartographers. Future studies could explore learning-based constraint modeling by integrating deep learning, visual perception models, or large-scale cartographic knowledge bases to automatically infer more comprehensive quality criteria.

Finally, the optimization efficiency remains dependent on the selected search strategy and parameter space definition. Although heuristic optimization methods provide effective solutions, large-scale production scenarios may require more efficient adaptive strategies. Future work could investigate reinforcement learning, surrogate optimization, or knowledge-guided parameter initialization to improve convergence efficiency and scalability for large-volume multiscale mapping applications.

## 6. Conclusion

This study presents a similarity-driven optimization framework for automated line generalization control. The framework formulates generalization as a constrained multiscale similarity optimization problem, where spatial similarity and cartographic constraints are jointly integrated into a unified objective function to determine appropriate algorithm parameters. Experiments on line simplification demonstrate that the proposed framework can effectively balance information preservation and cartographic requirements across different target scales. The results show that multiscale similarity can serve as an effective control objective, while cartographic constraints ensure readability, smoothness, and geometric validity during the optimization process. The framework also provides consistent parameter optimization across different simplification algorithms and optimization strategies. Overall, this study provides a quantitative and interpretable approach for adaptive generalization control by integrating similarity assessment, constraint modeling, and parameter optimization. Future work will extend the framework to multi-feature generalization and other cartographic transformation operations.

## Acknowledgments

## Disclosure statement

No potential conflict of interest was reported by the author(s).

## Notes on contributors

***Pengbo Li*** is Assistant Professor at the Faculty of Geomatics, Lanzhou Jiaotong University, Lanzhou, China. His research interests include map generalization, spatial relations and machine learning. He contributed to the concept, methodology, coding, dataset construction, and original manuscript writing.

***Haowen Yan*** is a Professor at the Faculty of Geomatics, Lanzhou Jiaotong University, Lanzhou, China. His research interests include map generalization, spatial relations, geovisualization, WeMaps and spatial data security. He contributed to the concept, funding acquisition, and manuscript revision.

***Xiaomin Lu*** is a Professor at the Faculty of Geomatics, Lanzhou Jiaotong University, Lanzhou, China. Her research interests include map generalization and spatial relations. In this paper, she contributed to preparing datasets and reviewing the manuscript.

***Binbin Lin*** is an Assistant Professor at the Faculty of Geomatics, Lanzhou Jiaotong University, Lanzhou, China. Her research interests include GeoAI, social sensing, and geospatial big data. In this paper, she contributed to preparing datasets and reviewing the manuscript.

## Funding

This work was supported by the National Natural Science Foundation of China [42501552, 42371463, 42471476]，Science-Technology Foundation for Young Scientists of Gansu Province [25JRRA202], Project supported by the Tianyou Postdoctoral Science Foundation of Lanzhou Jiaotong University [LJTYBH-2026017]